\documentclass[5p,final,authoryear,twocolumn]{elsarticle}
\usepackage{booktabs,float}
\usepackage{graphics}
\usepackage{hhline}
\usepackage{multirow}
\usepackage{graphicx}
\usepackage{amssymb}
\usepackage{mwe,graphbox,fixltx2e}
\usepackage[export]{adjustbox}
\usepackage{amsmath,textcomp,threeparttable,gensymb}
\usepackage{color,url,rotating}
\input{boldcommands}
\usepackage{tocloft}
\usepackage[bookmarks=true,bookmarksopen=false,
breaklinks=true,colorlinks=true,linkcolor=blue,anchorcolor=blue,citecolor=blue,
filecolor=blue,menucolor=blue,urlcolor=blue]{hyperref}

\newcommand{\ts}{\textsuperscript}
\begin{document}
\begin{frontmatter}

\title{Multiple Sclerosis Lesion Segmentation from Brain MRI via Fully
Convolutional Neural Networks}

\author[1]{Snehashis Roy\corref{cor1}}
\ead{snehashis.roy@nih.gov}
\author[1,2]{John A. Butman}
\author[2,3,4]{Daniel S. Reich}
\author[4]{Peter A. Calabresi}
\author[1]{Dzung L. Pham}
\address[1]{Center for Neuroscience and Regenerative Medicine, Henry M. Jackson Foundation, United 
States}
\address[2]{Radiology and Imaging Sciences, Clinical Center, National Institute of Health, 
United States}
\address[3]{Translational Neuroradiology Section, National Institute of Neurological Disorders and Stroke, United States}
\address[4]{Department of Neurology, Johns Hopkins University, United States}
\cortext[cor1]{Corresponding author}

\begin{abstract}
Multiple Sclerosis (MS) is an autoimmune disease that leads to lesions in the central nervous
system.  Magnetic resonance (MR) images provide sufficient imaging contrast to visualize and detect
lesions, particularly those in the white matter.  Quantitative measures based on various features of
lesions have been shown to be useful in clinical trials for evaluating therapies. Therefore robust
and accurate segmentation of white matter lesions from MR images can provide important information
about the disease status and progression.  In this paper, we propose a fully convolutional neural
network (CNN) based method to segment white matter lesions from multi-contrast MR images. The
proposed CNN based method contains two convolutional pathways. The first pathway consists of 
multiple parallel convolutional filter banks catering to multiple MR modalities. In the second 
pathway, the outputs of the first one are concatenated and another set of convolutional filters are
applied. The output of this last pathway produces a membership function for lesions that may be
thresholded to obtain a binary segmentation.  The proposed method is evaluated on a dataset of $100$
MS patients, as well as the \texttt{ISBI 2015} challenge data consisting of $14$ patients. 
The comparison is performed against four publicly available MS lesion 
segmentation methods. Significant improvement in segmentation quality over the 
competing methods is demonstrated on various metrics, such as Dice and 
false positive ratio. While evaluating on the \texttt{ISBI 2015} challenge data, our method 
produces a score of $90.48$, where a score of $90$ is considered to be comparable to a human rater.
\end{abstract}

\begin{keyword}
lesions \sep multiple sclerosis \sep CNN \sep deep learning \sep segmentation \sep neural networks \sep brain
\end{keyword}
\end{frontmatter}

\section{Introduction}
\label{sec:intro}
Multiple Sclerosis (MS) is an autoimmune disease of the central nervous system, in which 
inflammatory demyelination 
of axons causes focal lesions to occur in the brain.  White matter lesions in MS can be detected
with standard magnetic resonance imaging (MRI) acquisition protocols without contrast injection. It
has been shown that many features of lesions, such as volume \cite{kalincik2012} and location
\cite{sati2016}, are important biomarkers of MS, and can be used to detect disease onset or track
its progression. Therefore accurate segmentation of white matter lesions is important in
understanding the progression and prognosis of the disease. With $T_2$-w FLAIR (fluid attenuated
inversion recovery) imaging sequences, most lesions appear as bright regions in MR images, which
helps its automatic segmentation. Therefore FLAIR is the most common imaging contrast for detection
of MS lesions and is often used in conjunction with other structural MR contrasts, including
$T_1$-w, $T_2$-w, or $PD$-w images. Although manual delineations are considered as the gold
standard, manually segmenting lesions from 3D images is tedious, time consuming, and often not
reproducible. Therefore automated lesion segmentation from MRI is an active area of development in
MS research.

Automated lesion segmentation in MS is a challenging task for various reasons: (1) the 
lesions are highly variable in terms of size and location, (2) lesion boundaries are often not well 
defined, particularly on FLAIR images, and (3) clinical quality FLAIR images may possess low 
resolution and often have imaging artifacts. It has also been observed that there is very high 
inter-rater variability even with experienced raters \cite{carass2017,egger2017}. Therefore there 
is an inherent reliability challenge associated with lesion segmentation. This problem is 
accentuated by the fact that MRI does not have any uniform intensity scale (like CT); acquisition of 
images in different scanners and with different contrast properties can therefore add to the 
complexity of segmentation.

Many automated lesion segmentation methods have been proposed in the past decade 
\cite{lorenzo2013}.  There are usually two broad categories of segmentations, supervised and
unsupervised. Unsupervised lesion segmentation methods rely on intensity models of brain tissue, 
where image voxels containing high intensities in FLAIR images are modeled as outliers 
\cite{lorenzo2011,shiee2009} based on the intensity distributions. The outlier voxels then become 
potential candidates for lesions and then the segmentation can be refined by a simple threshold 
\cite{souplet2008,llado2015,jain2015}. Alternatively, Bayesian models such as mixtures
of Gaussians \cite{schmidt2012,strumia2016,leemput2001,sudre2015} or Student's t mixture models
\cite{ferrari2016} can be applied on the intensity distributions of potential lesions and normal
tissues. Optimal segmentation is then achieved via an expectation-maximization algorithm. Additional
information about intensity distributions and expected locations of normal tissues via a collection
of healthy subjects \cite{warfield2015} can be included to determine the lesions more accurately.
Local intensity information can also be included via Markov random field to obtain a smooth
segmentation \cite{harmouche2006,harmouche2015}. 

Supervised lesion segmentation methods make use of atlases or templates, which typically consist of
multi-contrast MR images and their manually delineated lesions. As seen in the 
\texttt{ISBI-2015}\footnote[1]{\url{https://smart-stats-tools.org/lesion-challenge-2015}} lesion 
segmentation
challenge \cite{carass2017}, supervised methods have become more popular and are usually superior
to unsupervised ones, with $4$ out of top $5$ methods being supervised.  These methods learn the
transformation from the MR image intensities to lesion labels (or memberships) on atlases, and then
the learnt transformation is applied onto a new unseen image to generate its lesion labels. Logistic
regression \cite{sweeney2013,sweeney2016} and support vector machines \cite{christos2008} have
been used in lesion classification, where features include voxel-wise intensities from
multi-contrast images and the classification task is to label an image voxel as lesion or
non-lesion. Instead of using voxel-wise intensities, patches have been shown to be a robust and
useful feature \cite{roy2014spie1}. Random forest \cite{maier2015,geremia2011,jog2015} and
k-nearest neighbors \cite{griffanti2016} based algorithms have used patches and other features,
computed at a particular voxel, to predict the label of that voxel.  Dictionary based methods
\cite{roy2015,roy2015mlmi,guizard2015,deshpande2015} use image patches from atlases to learn a
patch dictionary that can sufficiently describe potential lesion and non-lesion patches. For a new
unseen patch, similar patches are found from the dictionary and combined with weights based on the
similarity.

In recent years, convolutional neural networks (CNN), also known as deep learning 
\cite{hinton2015}, have been successfully applied to many medical image processing applications
\cite{summers2016,litjens2017}. CNN based methods produce state-of-the-art results in many computer 
vision problems such as object detection and recognition \cite{szegedy2015}. The primary advantage 
of neural networks over traditional machine learning algorithms is that CNNs do not need 
hand-crafted features, making it applicable to a diverse set of problems when it is not obvious what 
features are optimal. Because neural networks can handle 3D images or image patches, both 2D 
\cite{roth2016} and 3D \cite{brosch2016} algorithms have been proposed, with 2D 
patches often being preferred for memory and speed efficiency.  With advancements in graphics 
processor units (GPU), neural network models can be trained in a GPU within a fraction of time taken 
by that with multiple CPUs. Also CNNs can handle very large datasets without incurring too much 
increase in processing time. Therefore they have gained popularity in the medical imaging community 
in solving increasingly difficult problems.

CNNs have been shown to be better or on par with both probabilistic and multi-atlas label fusion
based methods for whole brain segmentation on adult \cite{wachinger2017,isgum2016a,chen2017} and
neonatal brains \cite{zhang2015,isgum2016b}. They have been especially successful in tumor
segmentations \cite{kamnitsas2017,pereira2015,veronica2016}, as seen on the \texttt{BRATS 2015}
challenge \cite{menze2015}.  They have recently been applied for brain extraction in the presence 
of tumors \cite{kleesiek2016}. Missing image contrasts pose a significant challenge in medical
imaging, where not all available image contrasts may not be acquired for all subjects. 
Traditional
CNN architectures can be modified to include image statistics in addition to image intensities to
circumvent missing image contrasts \cite{bengio2016} without sacrificing too much accuracy. CNN
models have also been applied to segment both cross-sectional 
\cite{prieto2017,yoo2014,ghafoorian2017a,ghafoorian2017b,moeskops2017} and longitudinal 
\cite{birenbaum2016} lesions from multi-contrast MR images. Recently, a two-step cascaded CNN 
architecture \cite{llado2017} has been
proposed, where two separate networks are learnt; the first one computes an initial lesion 
membership based on MR images and manual segmentations, while the second one refines the 
segmentation from the first network by including its false positives in the training samples.

In this paper, we propose a fully convolutional neural network model, called Fast Lesion EXtraction 
using COnvolutional Neural Networks (FLEXCONN), to segment MS lesions, 
where parallel pathways of convolutional filters are first applied to multiple contrasts.  The 
outputs of those pathways are then concatenated and another
set of convolutional filters is applied on the joined output. Similar to
\cite{ghafoorian2017b}, we used large 2D patches and show that larger patches produce more accurate 
results compared to smaller patches. The paper is organized
as follows. First the experimental data is described in Sec.~\ref{sec:materials}. The proposed 
FLEXCONN network architecture and its various parameter optimizations are described in 
Sec.~\ref{sec:method}. The segmentation results and the comparison with other methods are described 
in Sec.~\ref{sec:results}.

\begin{table}[!bt]
\caption{
Short description of the four datasets is presented here. Details can be found in Sec.~\ref{sec:materials}. For \texttt{ISBI-21} and \texttt{ISBI-61}, each image has two manual
lesion masks from two raters.
}
\tabcolsep 3pt
\begin{center}
\begin{tabular}{ccccc}
\toprule[2pt]
Dataset & \#Images & \#Masks & Usage & Availability \\
\cmidrule[2pt](lr){1-5}
\texttt{ISBI-21} & 21 & 42 & Training & Public \\
\texttt{VAL-28} & 28 & 28 & Validation & Private \\
\texttt{ISBI-61} & 61 & 122 & Testing & Private \\
\texttt{MS-100} & 100 & 100 & Testing & Private \\
\bottomrule[2pt]
\end{tabular}
\end{center}
\label{tab:dataset}
\end{table}

\begin{table}[!tb]
\caption{
Imaging parameters, such as repetition time $T_R$ (ms), echo time $T_E$(ms), inversion time $T_I$
(ms), flip angle, and resolution (mm\ts{3}) are shown. These parameters are same for all datasets 
described in Sec.~\ref{sec:materials}.
}

\tabcolsep 2pt
\begin{center}
\begin{tabular}{cccccc}
\toprule[2pt]
& $T_R$ & $T_E$ & $T_I$ & Flip & Resolution \\
& & & & Angle & \\
\cmidrule[2pt](lr){1-6}
{3D MPRAGE} & 10.3 & 6 & 835 & 8\degree & $0.82\times 0.82\times 1.17$\\
$T_2$-w & 4177 & 12.31 & N/A & 90\degree & $0.82\times 0.82\times 2.2$\\
$PD$-w & $4177$ & 80 & N/A & 90\degree & $0.82\times 0.82\times 2.2$\\
2D FLAIR & $11000$ & 68 & 2800 & 90\degree & $0.82\times 0.82\times 2.2$ \\
& & & & & \& $0.82\times 0.82\times 4.4$\\
\bottomrule[2pt]
\end{tabular}
\end{center}
\label{tab:scan_param}
\end{table}

\section{Materials}
\label{sec:materials}
Two sets of data are used to evaluate the proposed algorithm. The first dataset is from the 
\texttt{ISBI 2015} challenge \cite{carass2017}, which includes two groups, training and 
testing. The training group, denoted by \texttt{ISBI-21}, is publicly available and comprises 
$21$ scans from $5$ subjects. Four of the subjects have $4$ time-points and one has $5$ time-points, 
each time-point separated by approximately a year.  The test group, denoted by \texttt{ISBI-61}, is
not public and has $14$ subjects with $61$ images, each subject with $4-5$ time-points, each 
time-point also being separated by a year. Although these images actually contain longitudinal scans 
of the same subject, we treat the dataset as a cross-sectional study and report numbers on each 
image separately since longitudinal information is not used within our approach. 
A short description of the datasets is provided in Table~\ref{tab:dataset}.

The second dataset consists of $128$ patients enrolled in a natural history study of MS, $79$ with 
relapsing-remitting, $30$ with secondary progressive, and $19$ with primary progressive MS. For 
experimentation purpose, we arbitrarily divided this dataset into two groups, validation ($n=28$) 
and test ($n=100$), denoted as \texttt{VAL-28} and \texttt{MS-100} respectively. The proposed 
algorithm as well as the other competing methods were trained using \texttt{ISBI-21} as 
training data. Then various parameters, as described in Sec.~\ref{sec:param}, were optimized using 
\texttt{VAL-28} as the validation set. Finally the optimized algorithms were compared on the 
\texttt{ISBI-61} and \texttt{MS-100} datasets, as detailed in Sec.~\ref{sec:ms100} and 
Sec.~\ref{sec:isbi61}.

Each subject from both datasets had $T_1$-w MPRAGE, $T_2$-w, $PD$-w, and FLAIR images acquired in a
Philips 3T scanner. The imaging parameters are listed in Table~\ref{tab:scan_param}. Each image in 
\texttt{MS-100} and \texttt{VAL-28} has one manually delineated lesion segmentation mask. Every 
image in \texttt{ISBI-21} and \texttt{ISBI-61} has two masks, drawn by two different raters, as 
explained in \cite{carass2011}.

\section{Method}
\label{sec:method}

\subsection{Image Preprocessing}
\label{sec:preprocessing}
The $T_1$-w images of every subject in the \texttt{MS-100} and \texttt{VAL-28} dataset were first
rigidly registered \cite{avants2011} to the axial $1$ mm\ts{3} MNI template \cite{mori2009}. They 
were then skullstripped \cite{carass2011,roy2017} and corrected for any intensity inhomogeneity by 
N4 \cite{tustison2010}. The other contrasts, i.e. $T_2$-w, $PD$-w, and FLAIR images were then 
registered to the $T_1$-w image in MNI space, stripped with the same skull-stripping mask, and 
corrected by N4 after stripping.

The preprocessing steps for the \texttt{ISBI-21} and \texttt{ISBI-61} datasets were very similar 
and detailed in \cite{carass2017}. Briefly, the $T_1$-w images of the baseline of every
subject were rigidly registered to the MNI template, skullstripped \cite{carass2011}, and corrected
by N4. Then the other contrasts of the baseline and all contrasts of the followup time-points were
rigidly registered to the baseline $T_1$-w and corrected by N4.  Lesions for both data sets were
manually delineated on pre-processed FLAIR images, although the other contrasts were available for
reference.

\begin{figure*}[!tbh]
\begin{center}
\includegraphics[height=0.55\textwidth]{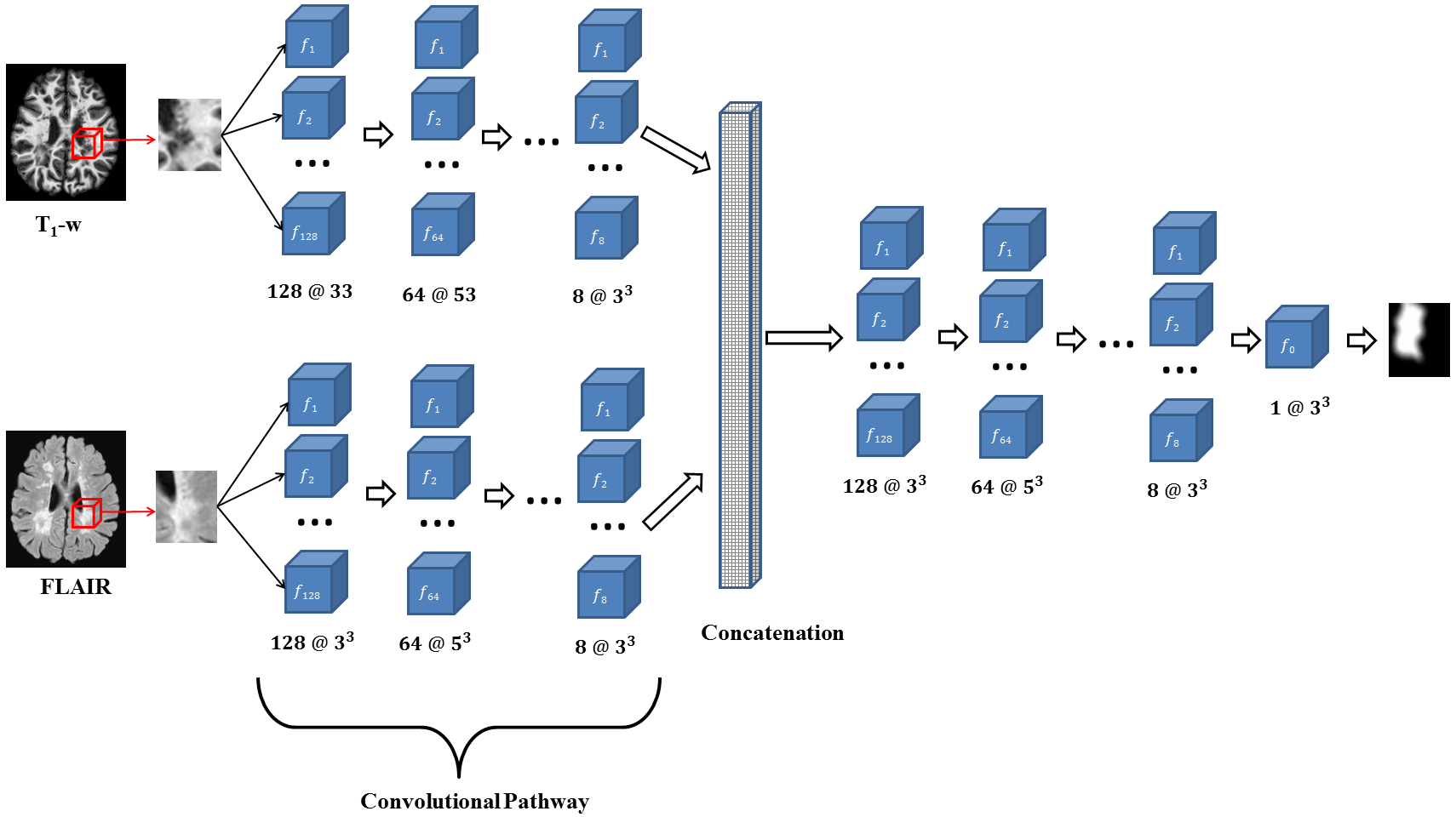}
\end{center}
\caption{Proposed neural network architecture for lesion segmentation is shown. 2D patches from 
multi-channel images are used as features  and convolutional filters ($f_j,j=1,2,\ldots$) are first 
applied in  parallel. Here $j$ denotes the index of the filter. Note that each ``filter ($f_j$)" 
includes a convolution and a ReLU module. The filter outputs are concatenated and passed  through 
another convolutional pathway to predict a membership function of the patch. The filter number and 
sizes are shown as $128\ @ \ 3^2$, indicating the corresponding filter bank contains $128$ filters, 
each with size $3\times 3$.  }
\label{fig:Fig1}
\end{figure*}

\subsection{CNN Architecture}
\label{sec:cnn}
Cascade type neural network architectures have become popular in medical image 
segmentation, where features are either 2D slices or 3D patches from MR images.  Typically, 
multi-channel patches are first independently passed through convolutional filter banks, then a 
fully connected (FC) layer is applied to predict the voxel-wise membership at the center of the 
patches \cite{wachinger2017} from the concatenated outputs of the filters. We follow a similar 
architecture, shown in Fig.~\ref{fig:Fig1}, where multi-channel 2D $p_1\times p_2$ patches are 
convolved with multiple filter banks of various sizes (called a ``convolutional 
pathway"), and the outputs of the convolutional pathways are concatenated. The details 
of a convolutional pathway is given in Table~\ref{tab:conv_architecture}.  After concatenation, 
instead of an FC layer to predict the membership or probability of the center voxel of the 
$p_1\times p_2$ patch, we add another convolutional pathway that predicts a membership value of the 
whole $p_1\times p_2$ patch. Note that with variable pad sizes (see 
Table~\ref{tab:conv_architecture}), the sizes of the input and outputs of the filters are kept 
identical to the original MR image patch size. The training memberships are generated by simply 
convolving the manual hard segmentations with a $3\times 3$ (denoted $3^2$) Gaussian kernel. We 
observed that larger patches produce mored accurate segmentations compared to smaller patches, and  
determined that a $35\times 35$ patch produced the best results based on the \texttt{VAL-28} 
dataset. The estimation of the optimal patch size is described in Sec.~\ref{sec:param}.

\begin{table*}[htb]
\caption{Filter parameters of a convolutional pathway, as shown in  Fig.~\ref{fig:Fig1}, are 
provided.  Every size is in voxels. Note that with variable pad sizes, the input and output size of 
the filters are kept identical to the training patch size $p_1\times  p_2$.}
\begin{center}
\begin{tabular}{clccccc}
\toprule[1pt]
Filter & Type & Number & Filter Size & Pad Size & Parameters & \# Parameters  \\
Bank &  & of Filters &  & & &   \\
\cmidrule[1.5pt](lr){1-7}%
1 & Convolution & 128 & $3^2$ & $1^2$ & $3\times 3\times 128$  & 1152 \\
2 & Convolution & 64  & $5^2$ & $2^2$ & $5\times 5\times 128 \times 64$ & 204800\\
3 & Convolution & 32  & $3^2$ & $1^2$ & $3\times 3\times 64 \times 32$ & 18432\\
4 & Convolution & 16  & $5^2$ & $2^2$ & $5\times 5\times 32\times 16$ & 12800 \\
5 & Convolution & 8   & $3^2$ & $1^2$ & $3\times 3\times 16 \times 8$ & 1152\\
\bottomrule[2pt]
\end{tabular}
\end{center}
\label{tab:conv_architecture}
\end{table*}

\begin{figure*}[!tbh]
\begin{center}
\tabcolsep 0pt
\begin{tabular}{ccccc}
\textbf{MPRAGE} & \textbf{FLAIR} & \textbf{Manual} & \textbf{With FC} & \textbf{Without FC} \\
\includegraphics[width=0.19\textwidth]{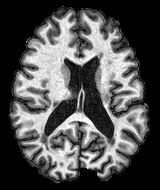} &
\includegraphics[width=0.19\textwidth]{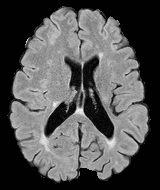} &
\includegraphics[width=0.19\textwidth]{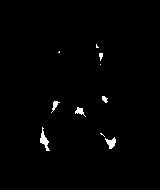} &
\includegraphics[width=0.19\textwidth]{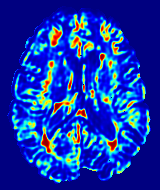} &
\includegraphics[width=0.19\textwidth]{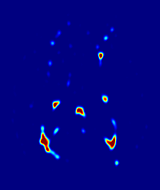} \\
\end{tabular}
\end{center}
\caption{Example of lesion memberships are shown when generated with a fully connected (FC) layer
predicting memberships at a voxel, compared to the proposed model where patch based memberships
are predicted using convolutional pathways. Note that a FC layer produces fuzzier memberships and 
potentially more false positives. Memberships are scaled between $0$ (dark blue) and $1$ (red).
}
\label{fig:Fig2}
\end{figure*}

Improved segmentation results were achieved using a set of $5$ convolutional filter banks with
decreasing numbers of filters in one convolutional pathway, as shown in Fig.~\ref{fig:Fig1} and
Table~\ref{tab:conv_architecture}. The optimal number of filter banks in a pathway was also 
estimated from a validation strategy discussed in Sec.~\ref{sec:param}. Each convolution is followed 
by a rectified linear unit (ReLU) \cite{hinton2010}. The combination of convolution and ReLU is
indicated by $f_j$ in Fig.~\ref{fig:Fig1}. Our experiments showed that smaller filter sizes such as
$3^2$ and $5^2$ generally produce better segmentation than bigger filters (such as $7^2$ and $9^2$),
which was also observed before \cite{karen2015}. We hypothesize that since lesion boundaries are
often not well defined, small filters tend to capture the boundaries better. Also the number of free
parameters ($9$ for $3^2$) increases for larger filters ($49$ for $7^2$), which in
turn can either decrease the stability of the result or incur overfitting. However, smaller filters
may perform worse for larger lesions. Therefore we empirically used a combination of $3^2$ and $5^2$
filters based on our validation set \texttt{VAL-28}.

As noted, a major difference in the network architecture proposed here 
in contrast to other popular CNN based segmentation methods is the use of a convolutional 
layer to predict membership functions. The advantages of such a configuration compared to a FC layer 
are as follows:
\begin{enumerate}
\item Depending on the number of convolutions and the patch size, the number of 
free parameters for a FC layer can be large, thereby increasing the possibility of overfitting.
Recent successful deep learning networks such as ResNet \cite{he2016b} and GoogLeNet 
\cite{szegedy2015} have put more focus on fully convolutional networks and networks, with ResNet 
having no FC layer at all. Although dropout \cite{srivastava2014} has been proposed to reduce the 
effect of overfitting a network to the training data, the mechanism of randomly turning off
different neurons inherently results in slightly different segmentations every time the training is 
performed even with the same training data.

\item We observed that memberships predicted with an FC layer result 
in more false positives compared to a fully convolutional network. An example is shown in 
Fig.~\ref{fig:Fig2},  where lesion memberships are generated from MPRAGE and FLAIR using
the proposed model of convolutional pathways and a comparable model where the last convolutional 
pathway after concatenation (see Fig.~\ref{fig:Fig1}) is replaced with a FC layer predicting 
voxel-wise memberships. The membership image
generated with an FC layer, although being close to $1$ inside the lesions, has high values ($\ge 
0.5$) in the left and right frontal cortex where the FLAIR image shows some artifacts. However, the 
membership obtained with the proposed method shows relatively low values near the frontal cortex.

\item With FC layer, voxel-wise predictions are performed for each voxel on a new image. Therefore 
the prediction time for the whole image comprising millions of voxels can take some time even on a 
GPU, as mentioned in \cite{wachinger2017}. In contrast, with fully 
convolutional prediction, lesion membership estimation of a $1$ mm\ts{3} MR volume of size 
$181\times 217\times 181$ takes only a couple of seconds. Note that although patches are used for 
training, the final trained model contains only convolution filters and does not depend in any way 
on the input patch size. Therefore during testing, the lesion membership of a whole 2D slice, 
irrespective of the slice size, is predicted at a time  by applying convolutions on the whole slice. 
Without an FC layer, the images need not be decomposed into sub-regions, e.g., 
\cite{kamnitsas2017}. Consequently, there is no need to employ membership smoothing between 
sub-regions. In addition, since the training memberships, generated by Gaussian blurring of hard 
segmentations, are smooth, the resultant predicted memberships are also smooth (Fig.~\ref{fig:Fig2} 
last column).  

\end{enumerate}

MS lesions are heavily under-represented as a tissue class in a brain MR image, 
compared to GM or WM. In the training dataset
\texttt{ISBI-21}, lesions represent on an average $1$\% of all brain tissue voxels. For a binary 
lesion classification, most supervised machine learning algorithms thus require balanced 
training data \cite{he2009}, where number of patches with lesions are approximately equal to lesion 
free patches. Therefore normal tissue patches are randomly undersampled \cite{llado2017,roy2015}
to generate a balanced training dataset. This is true for a small $5^2$ or $7^2$ patch, which may 
have all or most voxels as lesions, thereby requiring some other patches with all or most voxels as 
normal tissue. In Sec.~\ref{sec:param}, we show that using larger patches, such as $25\times 25$ or 
$35\times 35$, produce more accurate segmentations compared to smaller $9^2$ or $13^2$ patches. 
Since we use large patches which cover most of the largest lesions, the effect of data imbalance 
is reduced.

With large patches, our training data consists of patches where the center  voxel 
of a patch has a lesion label, i.e., all lesion patches are included in the training data with a 
stride of $1$. We do not include any normal tissue patches, where none of the voxels have a lesion
label. Experiments showed that inclusion of the normal tissue patches does not improve segmentation 
accuracy, but incurs longer training time by requiring more training epochs to achieve similar
accuracy. However, one drawback of only including patches with lesions is that generally more 
training data are required, especially when the number of lesions become much less than number of 
parameters to be optimized, as shown in Table~\ref{tab:conv_architecture}. 

\begin{figure}[!tb]
\begin{center}
\includegraphics[width=0.47\textwidth]{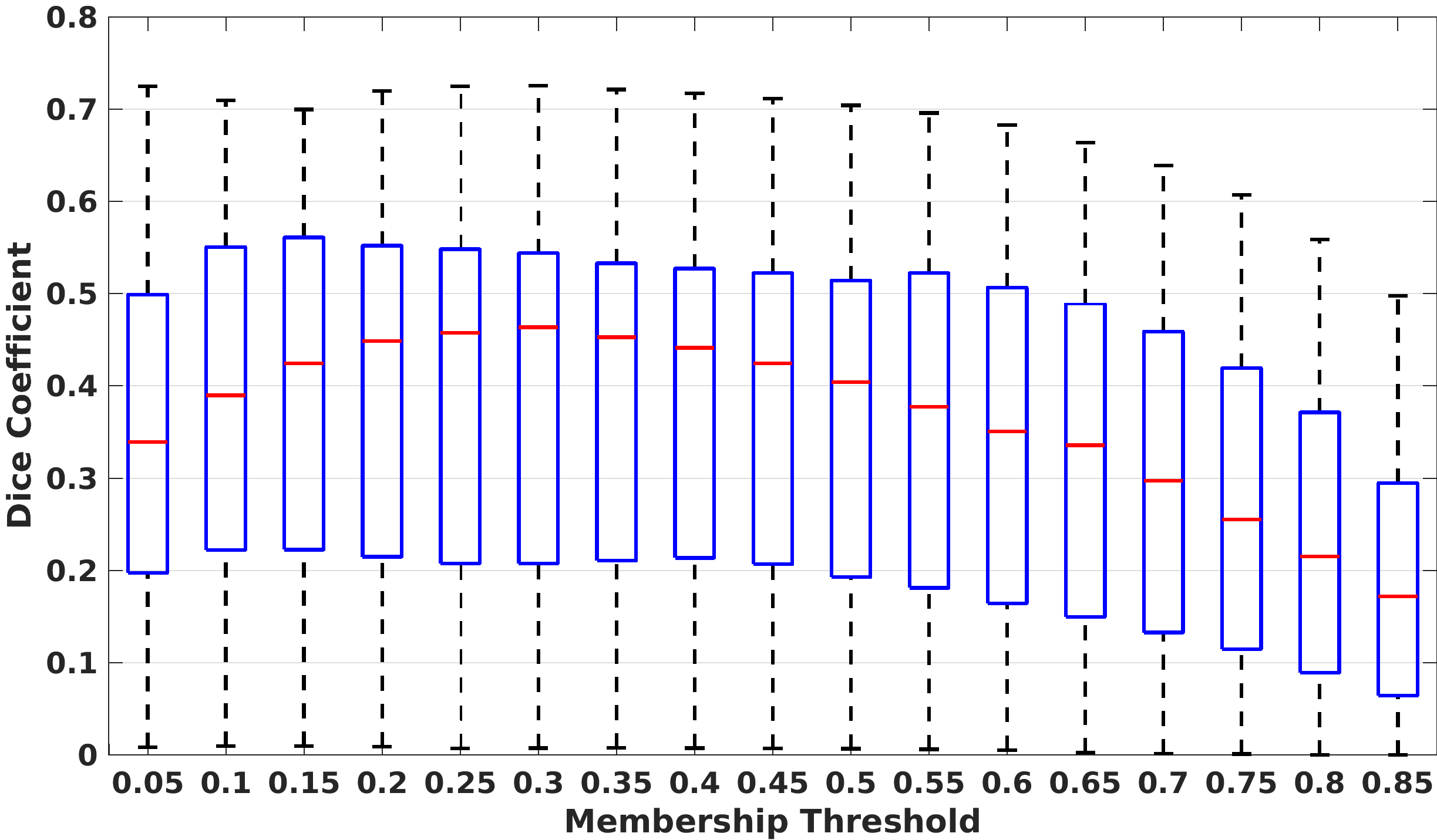} 
\end{center}
\caption{
Dice coefficients of segmentations from \texttt{VAL-28} dataset are shown at different
membership thresholds from $0.05$ to $0.85$. The highest median Dice was observed at $0.30$.  See
Sec.~\ref{sec:param} for details.  
}
\label{fig:Fig3}
\end{figure}


\begin{figure*}[!bth]
\begin{center}
\tabcolsep 2pt
\begin{tabular}{ccccc}
\texttt{MPRAGE} & \texttt{T2} & \texttt{PD} & \texttt{FLAIR} & \texttt{Manual} \\
\includegraphics[width=0.18\textwidth]{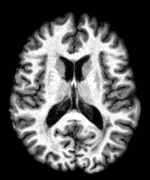} &
\includegraphics[width=0.18\textwidth]{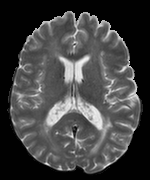} &
\includegraphics[width=0.18\textwidth]{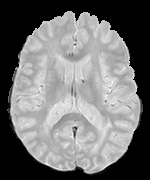} &
\includegraphics[width=0.18\textwidth]{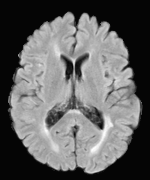} &
\includegraphics[width=0.18\textwidth]{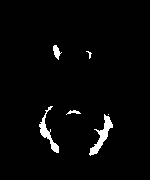} \\
\texttt{$13\times 13$} & \texttt{$17\times 17$} & \texttt{$19\times 19$} 
& \texttt{$31\times 31$} & \texttt{$35\times 35$} \\
\includegraphics[width=0.18\textwidth]{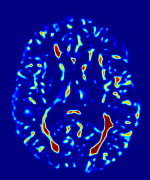} &
\includegraphics[width=0.18\textwidth]{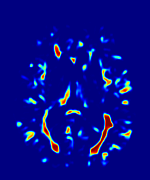} &
\includegraphics[width=0.18\textwidth]{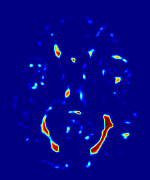} &
\includegraphics[width=0.18\textwidth]{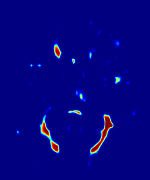} &
\includegraphics[width=0.18\textwidth]{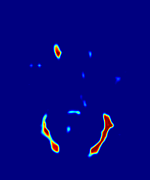} 
\end{tabular}
\end{center}
\caption{Memberships of a subject from \texttt{VAL-28} dataset with various patch sizes are shown.
As the patch size increases, the false positives in the cortex begins to decrease.
}
\label{fig:diff_patch_size}
\end{figure*}

\subsection{Comparison Metrics}
\label{sec:metrics}
We chose $4$ comparison metrics: Dice coefficient, lesion false positive rate (LFPR), positive 
predictive value (PPV), and absolute volume difference (VD) to compare segmentations. For a manual 
and an automated binary segmentation $\mathcal{M}$ and $\mathcal{A}$ 
respectively, Dice is a voxel-wise overlap measure defined as,
\begin{eqnarray*}
\mathrm{Dice}(\mathcal{A},\mathcal{M})=\frac{2|\mathcal{A} 
\cap  \mathcal{M}|}{|\mathcal{A}|+|\mathcal{M}|},
\end{eqnarray*}
where $|\cdot|$ denotes number of non-zero voxels. Since lesions are often small and their total 
volumes are typically very small ($1-2$\%) compared to the whole brain volume, Dice can be affected 
by the low volume of the segmentations \cite{geremia2011}. Therefore LFPR is defined based on 
distinct lesion counts. A distinct lesion is defined as an $18$-connected object,
although such a description of lesions may or may not be biologically accurate.
LFPR is the number of lesions in the automated segmentation that do not overlap with any lesions in 
the manual segmentation, divided by the total number of lesions in the automated segmentation. Two 
lesions are considered overlapped when they share at least one voxel. PPV is defined as the ratio of 
true positive voxels and total number of positive voxels, expressed as
\begin{eqnarray*}
\mathrm{PPV}(\mathcal{A},\mathcal{M})=\frac{2|\mathcal{A} \cap  \mathcal{M}|}{|\mathcal{A}|},
\end{eqnarray*}
Absolute volume difference is defined as
\begin{eqnarray*}
\mathrm{VD}(\mathcal{A},\mathcal{M}) = 
\frac{\mathrm{abs}(|\mathcal{A}| - |\mathcal{M}|)}{|\mathcal{M}|}.
\end{eqnarray*}
All statistical tests were performed with a non-parametric paired Wilcoxon signed rank test.

\subsection{Parameter Optimization}
\label{sec:param}
In this section, we describe a validation strategy to optimize user selectable parameters of the 
proposed network: (1) patch size, (2) number of filter banks in a convolutional pathway, and (3) the 
final threshold to create hard segmentations from memberships. After training with \texttt{ISBI-21}, 
the network was applied to the images of \texttt{VAL-28} to generate their lesion membership images. 
Memberships were thresholded and then masked with a cerebral white matter mask \cite{shiee2009} to  
remove any residual false positives. Dice was used as the primary metric for optimizing the 
parameters, with LFPR used as a secondary metric for patch size optimization. Although our model is 
capable of using all four available contrasts, initial experiments on \texttt{VAL-28} data showed 
negligible improvement in segmentation accuracy with $T_2$-w and $PD$-w images. Therefore all 
results were obtained with only MPRAGE and FLAIR contrasts.

\begin{figure*}[tbh]
\begin{center}
\includegraphics[width=0.6\textwidth]{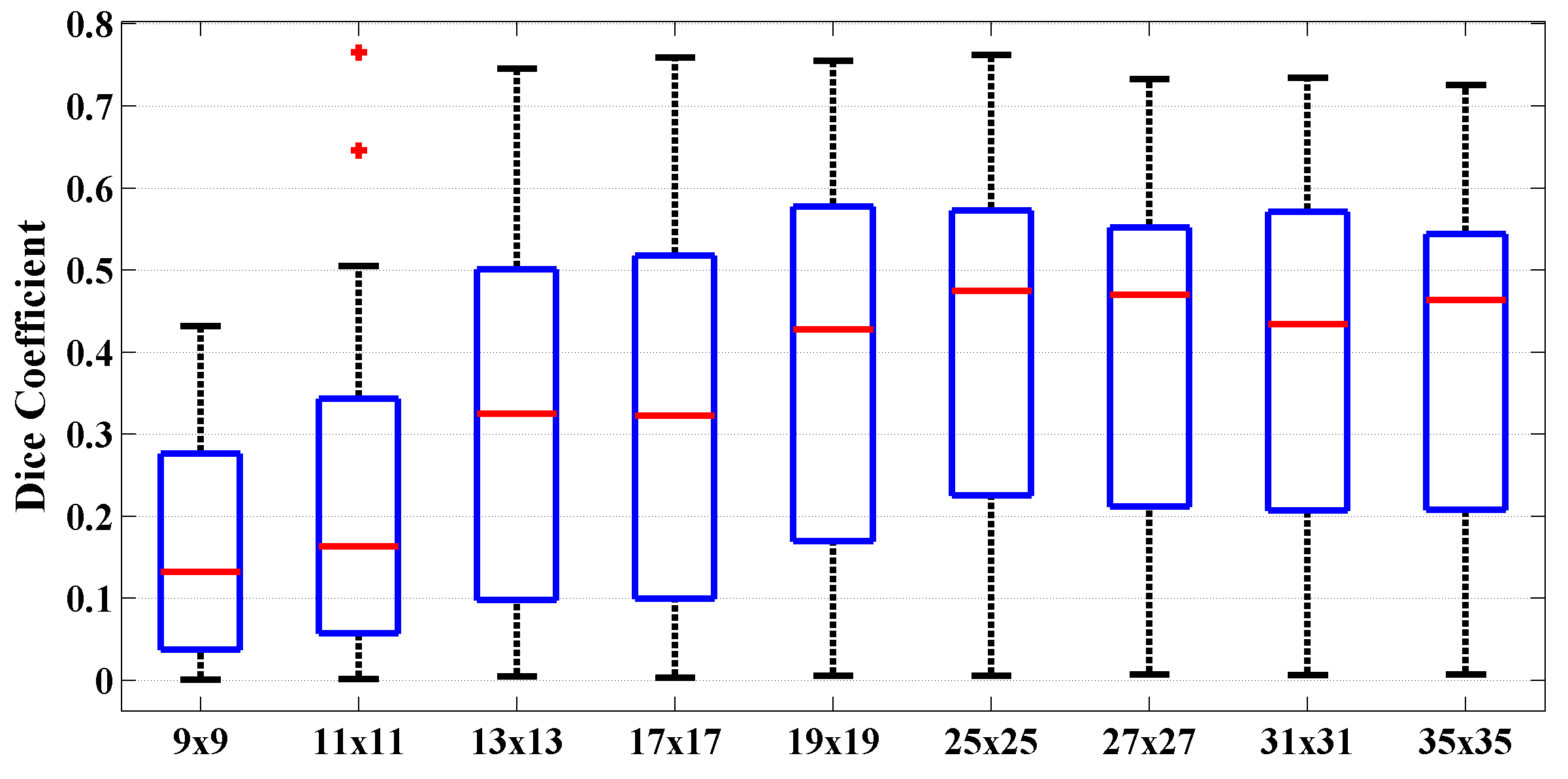} \\
\includegraphics[width=0.6\textwidth]{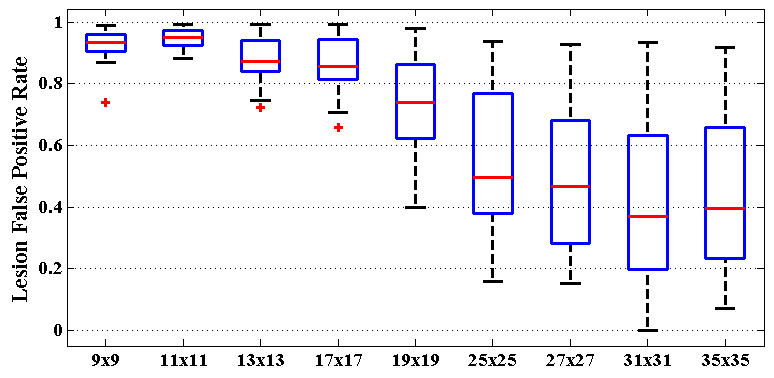} 
\end{center}
\caption{Dice coefficients (top) and LFPR (bottom) are plotted for the segmentations of 
\texttt{VAL-28} data when trained with \texttt{ISBI-21} for various patch sizes. See 
Sec.~\ref{sec:param} for details.
}
\label{fig:Fig5}
\end{figure*}

To optimize the membership threshold, we trained a network with $35\times 35$ patches. Memberships 
generated on \texttt{VAL-28} were segmented with thresholds from $0.05$ to $0.85$ with an increment 
of $0.05$. The range of Dice coefficients is shown in Fig.~\ref{fig:Fig3}. 
The highest median Dice coefficient was observed for a threshold of
$0.30$. This is intuitively reasonable because during training, the lesion memberships of atlases
were generated from their hard segmentations using a $3\times 3$ Gaussian kernel, and it can 
be shown that the half max of a $3\times 3$ discrete Gaussian is at $0.31$.

Next we varied the depth of a convolutional pathway from $2$ to $6$ filter banks while keeping
the number of filters as a multiple of $2$, with the last filter bank having $8$ filters. 
The highest median Dice coefficient was observed at a depth of $5$, which is
significantly larger than Dice coefficients with depths $3$ and $4$ ($p<0.05$). Although the 
differences in Dice coefficients were small between various depths, we used a depth of $5$ for the 
rest of the experiments. With more than $6$ filter banks, the Dice slowly decreases, which can be 
attributed to overfitting the training data.

Patch size is another important parameter of the network. In computer vision applications such as 
object detection, usually a whole 2D image is used as a feature. However, full 3D medical images can 
not typically be used because of memory limitations. Fig.~\ref{fig:diff_patch_size} shows examples 
of lesion memberships obtained with different sized 2D patches. As the patch sizes increases, the 
false positives that are mostly observed in the cortex tend to decrease. 
Fig.~\ref{fig:Fig5} shows a plot of Dice and LFPR with various patch sizes, ordered from left to 
right according to their increasing size. Note that smaller patches ($9^2$ to $17^2$) produced 
significantly lower Dice and higher LFPR compared to other patches ($p<0.001$), as seen from the 
memberships in Fig.~\ref{fig:diff_patch_size}. Also some of the highest Dice and lowest LFPR were 
observed for patches with large in-plane size, i.e., $31\times 31$, $27\times 27$, and $35\times 
35$. It was observed in Fig.~\ref{fig:Fig5} that there is no significant difference between Dice 
coefficients for $31\times 
31$, $35\times 35$, or $27\times 27$, but LFPR of both $35\times 35$ and $31\times 31$ are 
significantly lower than that of $27 \times 27$ ($p<0.05$). We chose $35\times 35$ as the optimal 
patch size. Other choices of smaller $5^2$ and $7^2$ patches (not shown) yielded worse results. 
Note that although training was performed with different patch sizes, the
memberships were generated slice by slice, as the trained model consisted only of convolutions
and did not need any information about patch sizes.


\subsection{Competing Methods}
\label{sec:competing_method}
We compared FLEXCONN with LesionTOADS \cite{shiee2009}, OASIS \cite{sweeney2013},
LST \cite{schmidt2012}, and S3DL \cite{roy2015}. LesionTOADS (Topology reserving Anatomy Driven 
Segmentation) does not need any parameter tuning and uses MPRAGE and FLAIR. OASIS (OASIS is 
Automated Statistical Inference for Segmentation) has a threshold parameter that is used to 
threshold the memberships to create a hard segmentation. It was optimized as 
$0.15$ by training a logistic regression on the \texttt{ISBI-21} and applying the regression model 
to \texttt{VAL-28}. A similar value was reported in the original paper. OASIS requires all four
contrasts, MPRAGE, $T_2$-w, $PD$-w, and FLAIR.  LST (Lesion Segmentation Toolbox) has a parameter 
$\kappa$, which initializes the lesion segmentation. Lower values of $\kappa$ produces bigger 
lesions. We optimized $\kappa$ to maximize the Dice coefficient on \texttt{VAL-28} data and found 
that $\kappa=0.10$ yielded the highest median Dice. LST uses MPRAGE and FLAIR images. S3DL has two 
parameters, number of atlases and membership threshold. We observed that adding more than $4$ 
atlases did not improve Dice coefficients significantly, as was reported in the original paper. 
Hence we used $5$ atlases as the last time-points of the $5$ subjects from the 
\texttt{ISBI-21} dataset. The optimal threshold for  S3DL was also found to be $0.80$. S3DL used 
MPRAGE and FLAIR as adding $T_2$-w and $PD$-w images did not improve the segmentation.

\subsection{Implementation Details}
\label{sec:implementation}
Our model was implemented in Tensorflow and Keras\footnote[7]{\url{https://keras.io/}}.  We used 
Adam \cite{kingma2015} as the optimizer, which has been shown to produce the fastest convergence in 
neural network parameter optimization. The optimization was run with fixed learning rate of $0.0001$ 
for $20$ epochs, which was empirically found to produce sufficient convergence without overfitting. 
During training with $35\times 35$ patches using lesions from $21$ subjects of \texttt{ISBI-21} 
dataset, we used $20$\% of the total number patches for validation and the remaining $80$\% for 
training. Training with $128$ minibatches required about $6$ hours on an Nvidia Titan X GPU
with $12$ GB memory. Segmenting a new subject took $3-5$ seconds.

\section{Results}
\label{sec:results}
In this section, we show comparison of FLEXCONN with other methods on two datasets \texttt{MS-100} 
and \texttt{ISBI-61} (see Section~\ref{sec:materials}). Research 
code\footnote[6]{\url{http://www.nitrc.org/projects/flexconn}} implementing our method is freely
available.

\subsection{\texttt{MS-100} Dataset}
\label{sec:ms100}
For this dataset, the training was performed separately with two sets of masks from the two raters 
of \texttt{ISBI-21} data. Then two memberships were generated for each of the $100$ images. For each 
image, the two memberships were averaged and thresholded to form the final segmentation.  
Fig.~\ref{fig:ms100-example} shows MR images and segmentations of $3$ subjects from the  
\texttt{MS-100} dataset, where the subjects have high ($22$cc), moderate ($8$cc), and low ($1$cc) 
lesion loads. For the subject with high lesion loads (\#1), all $5$ methods performed comparably, 
although OASIS and LST underestimated some small and subtle lesions (yellow arrow). For the subject 
with moderate lesion load (\#2), OASIS and S3DL underestimated some lesions (orange arrow) and 
LesionTOADS overestimated some (green arrow). When the lesion load is small and the FLAIR image has 
some artifacts (subject \#3), LesionTOADS, S3DL, and OASIS produce a false positive (yellow arrow) 
in the cortex. LST shows 
underestimation, but FLEXCONN does not produce the false positive. The reason is partly because of 
the use of large patches, which can successfully distinguish between bright voxels in cortex and 
peri-ventricular regions.

\begin{figure*}[!tbh]
\begin{center}
\tabcolsep 0pt
\begin{tabular}{ccccc}
\textsc{LesionTOADS} & \textsc{S3DL} & \textsc{OASIS} & \textsc{LST} & \textsc{Proposed} \\
\includegraphics[width=0.2\textwidth]{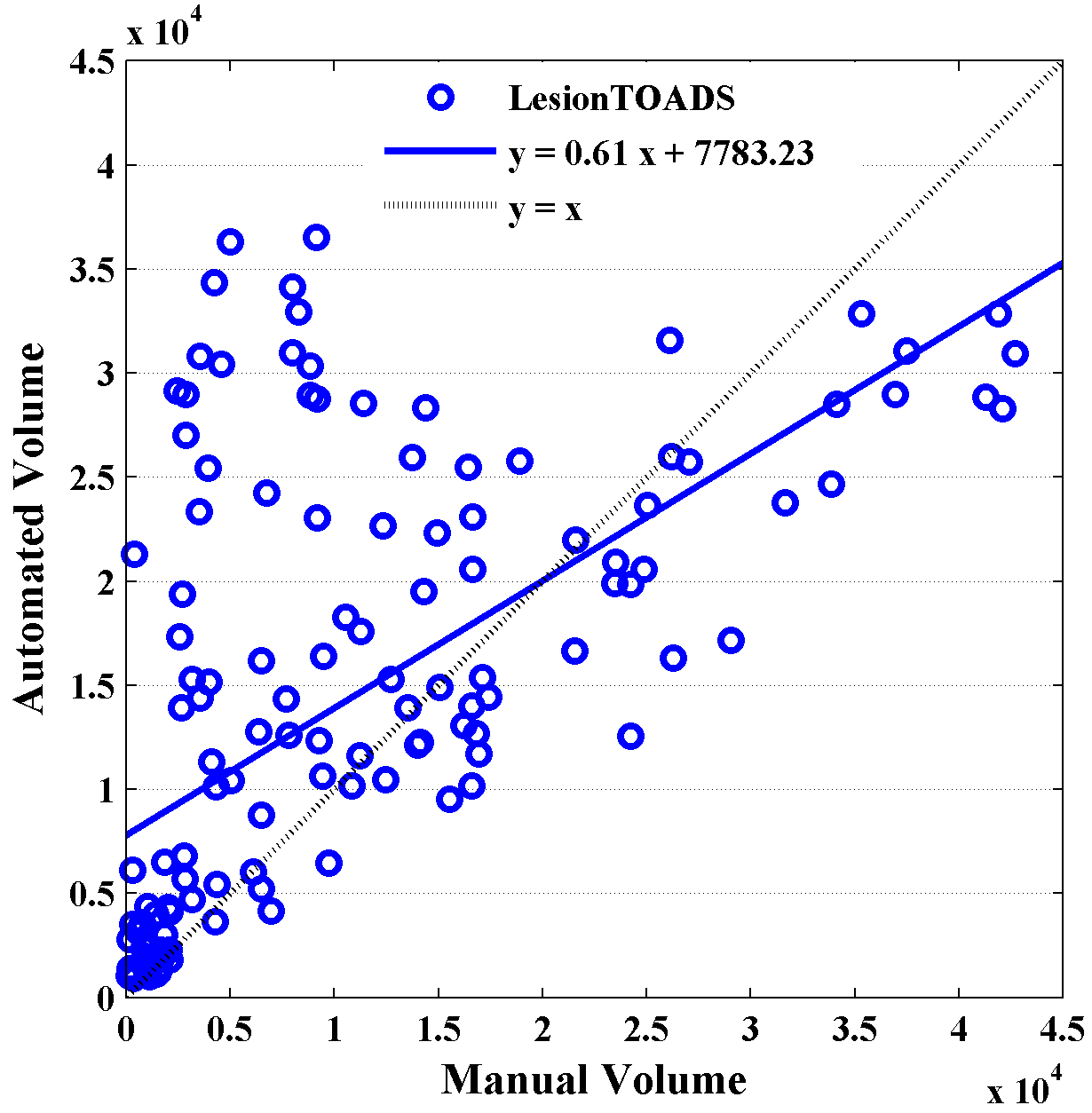} &
\includegraphics[width=0.2\textwidth]{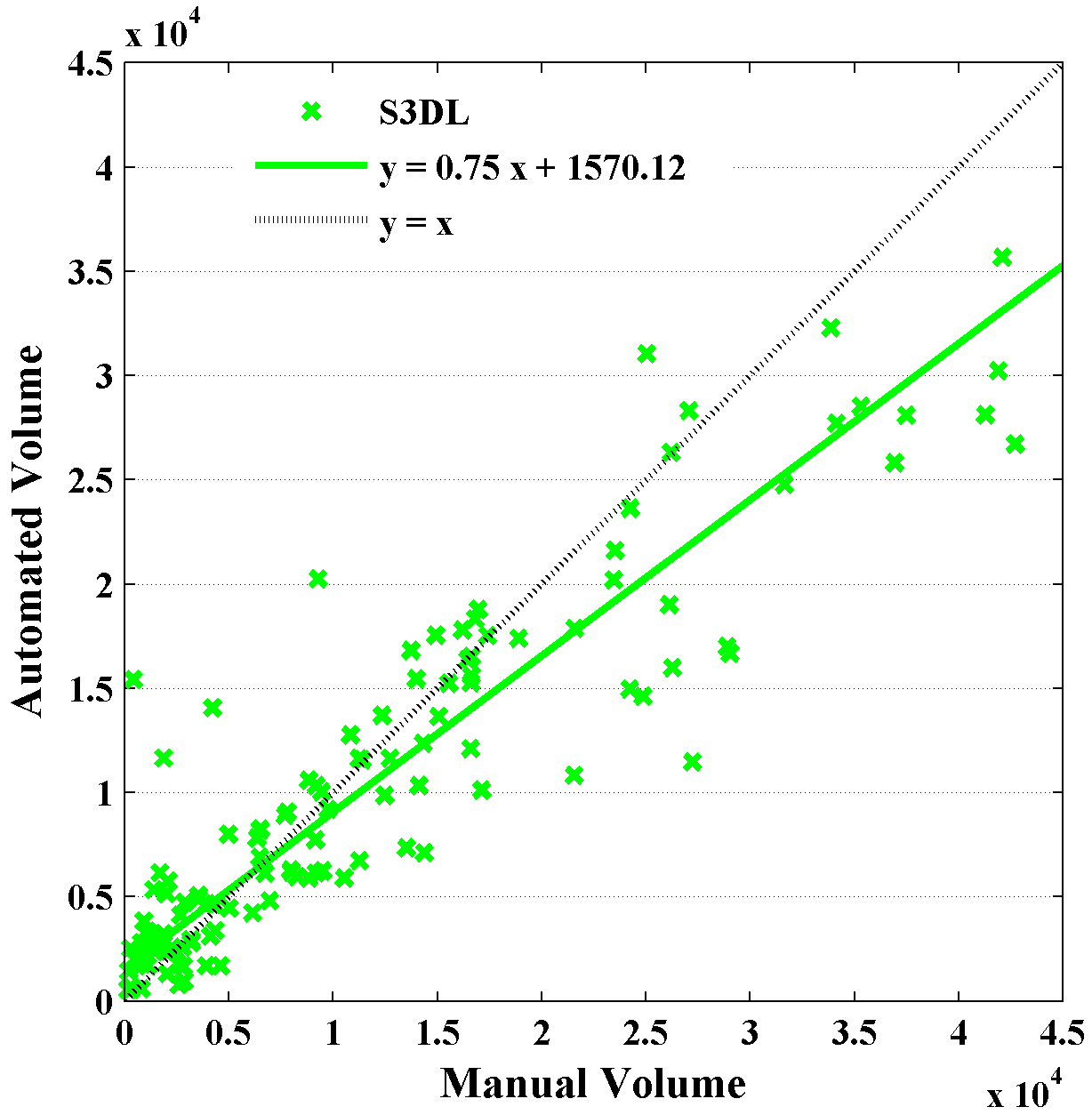} &
\includegraphics[width=0.2\textwidth]{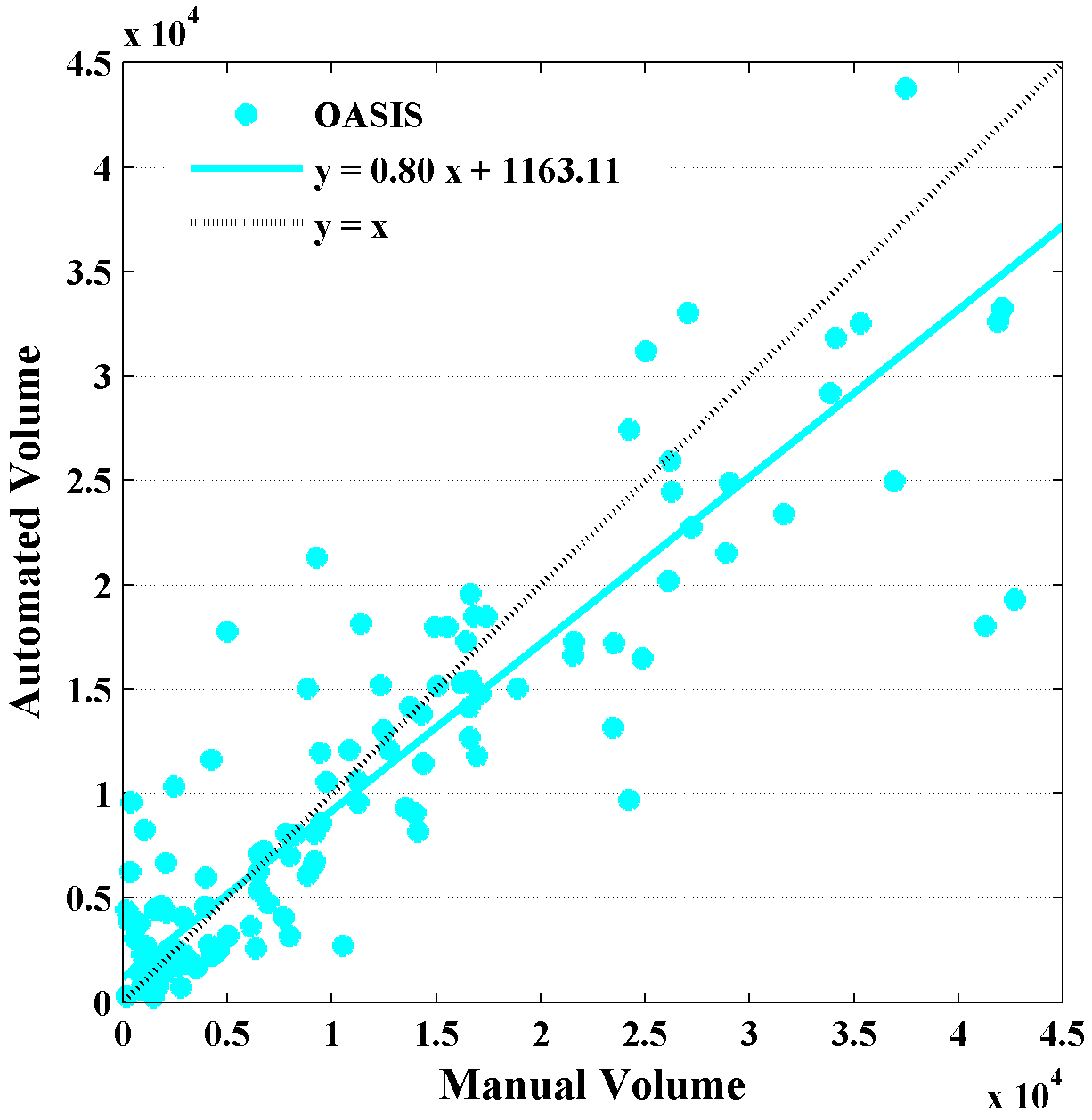} &
\includegraphics[width=0.2\textwidth]{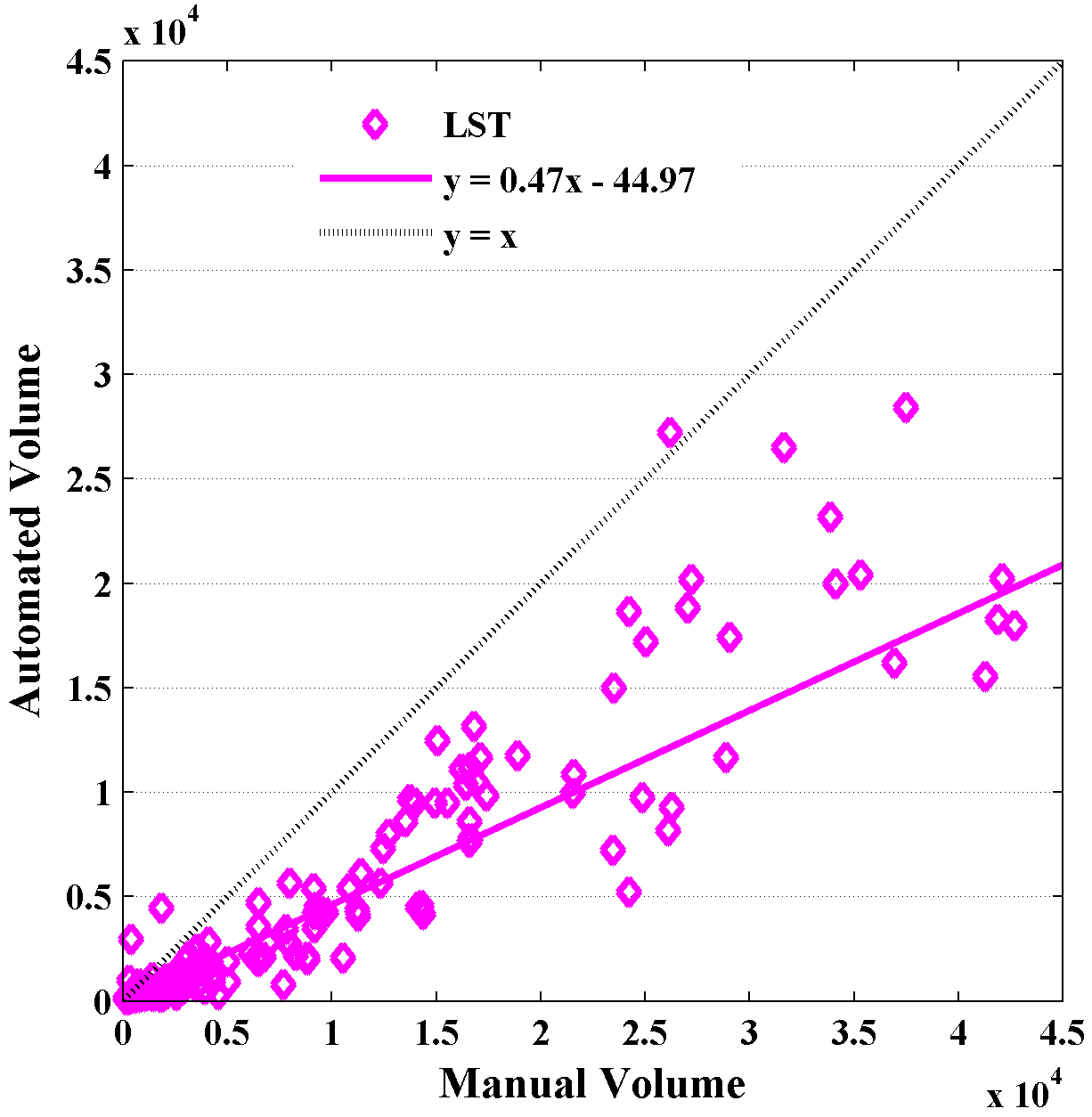} &
\includegraphics[width=0.2\textwidth]{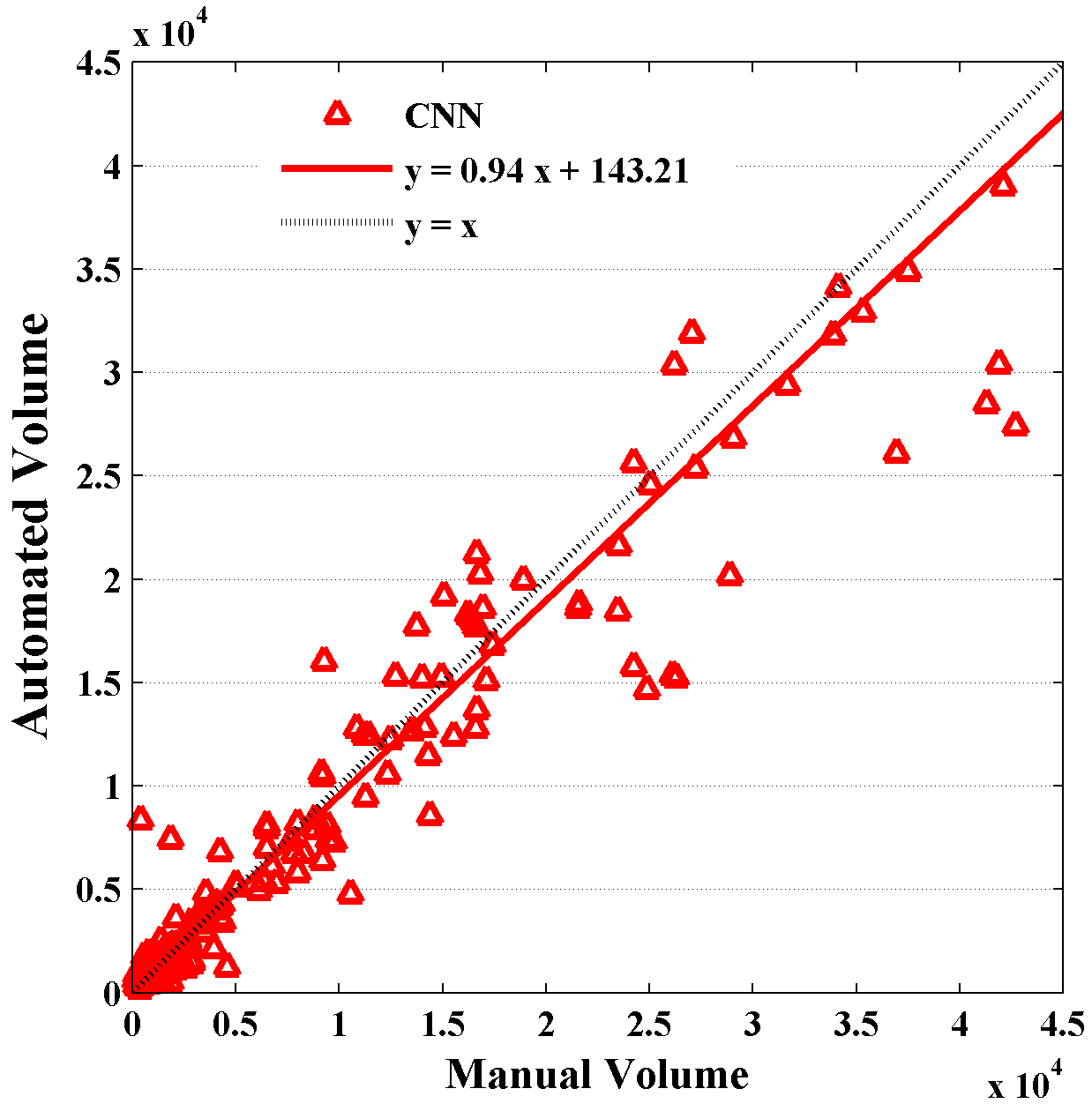} 
\end{tabular}
\end{center}
Manual vs automated lesion volumes for the $5$ methods on \texttt{MS-100} dataset. The solid lines 
show robust linear fits of the points and the dotted black line represents the unit slope line. 
Numbers are in mm\ts{3}.
\label{fig:ms100-scatterplot}
\end{figure*}

\begin{table*}[!tbh]
\caption{Slopes and intercepts from Fig.~\ref{fig:ms100-scatterplot} are shown for the 
\texttt{MS-100} dataset. Bold indicates the largest absolute value among the $5$ methods.}
\begin{center}
\tabcolsep 5pt
\begin{tabular}{ccc}
\toprule[2pt]
Method & Slope & Intercept (cc) \\
\cmidrule[2pt](lr){1-3}%
LesionTOADS \cite{shiee2009} & 0.6112 & 7783.2 \\
S3DL \cite{roy2015} & 0.7488 & 1570.1 \\
OASIS \cite{sweeney2013} & 0.8002 & 1163.1\\
LST \cite{schmidt2012} & 0.4650 & \textbf{-44.9} \\
FLEXCONN & \textbf{0.9421} & 143.2 \\
\bottomrule[2pt]
\end{tabular}
\end{center}
\label{tab:ms100-stats}
\end{table*}

\begin{table*}[!tbh]
Median values of Dice, lesion false positive rate (LFPR), positive predictive value (PPV), and 
volume difference (VD) are shown for competing methods on \texttt{MS-100} dataset. Bold indicates 
significantly highest/lowest $(p<0.05)$ number. See Sec.~\ref{sec:metrics} for the definition of the 
metrics.
\begin{center}
\tabcolsep 5pt
\begin{tabular}{ccccc}
\toprule[2pt]
& Dice & LFPR & PPV & VD \\
\cmidrule[1pt](lr){1-5}
LesionTOADS & 0.4678 & 0.6865 & 0.3968 & 0.4718 \\
S3DL        & 0.5526 & 0.4164 & 0.5968 & 0.2755 \\
OASIS       & 0.4993 & 0.5081 & 0.6242 & 0.2681 \\
LST         & 0.4239 & 0.4409 & \textbf{0.7820} & 0.5623 \\
FLEXCONN    & \textbf{0.5639} & \textbf{0.3077} & 0.6040 & \textbf{0.1978} \\
\bottomrule[1pt]
\end{tabular}
\end{center}
\label{tab:ms100-volstats}
\end{table*}

\begin{figure*}[!tbh]
\begin{center}
\tabcolsep 1pt
\begin{tabular}{cccccc}
& \textbf{MPRAGE} & \textbf{FLAIR} & $\mathbf{T_2}$-w & $\mathbf{PD}$-w & \textbf{Manual} \\
{\rotatebox{90}{\hspace{1em}\texttt{Subject \#1}}} & 
\includegraphics[width=0.13\textwidth]{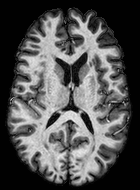} &
\includegraphics[width=0.13\textwidth]{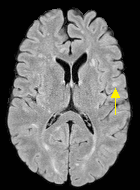} &
\includegraphics[width=0.13\textwidth]{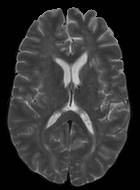} &
\includegraphics[width=0.13\textwidth]{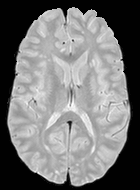} &
\includegraphics[width=0.13\textwidth]{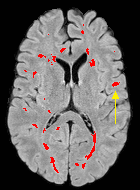} 
\end{tabular}
\tabcolsep 1pt
\begin{tabular}{cccccc}
\textbf{LesionTOADS} & \textbf{S3DL} & \textbf{OASIS} & \textbf{LST} & \textbf{FLEXCONN} & \textbf{Membership} \\
\includegraphics[width=0.13\textwidth]{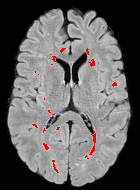} &
\includegraphics[width=0.13\textwidth]{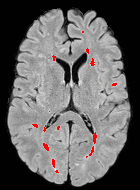} &
\includegraphics[width=0.13\textwidth]{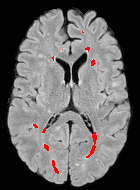} &
\includegraphics[width=0.13\textwidth]{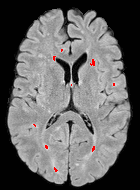} &
\includegraphics[width=0.13\textwidth]{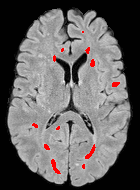} &
\includegraphics[width=0.13\textwidth]{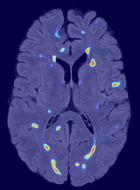} 
\end{tabular}
\tabcolsep 1pt
\begin{tabular}{cccccc}
& \textbf{MPRAGE} & \textbf{FLAIR} & $\mathbf{T_2}$-w & $\mathbf{PD}$-w & \textbf{Manual} \\
{\rotatebox{90}{\hspace{1em}\texttt{Subject \#2}}} & 
\includegraphics[width=0.13\textwidth]{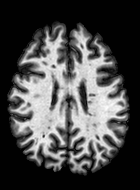} &
\includegraphics[width=0.13\textwidth]{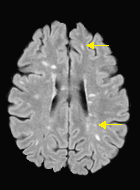} &
\includegraphics[width=0.13\textwidth]{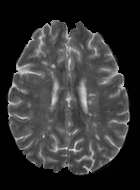} &
\includegraphics[width=0.13\textwidth]{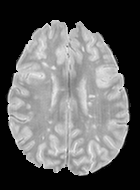} &
\includegraphics[width=0.13\textwidth]{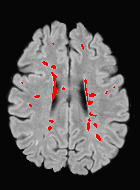} 
\end{tabular}
\tabcolsep 1pt
\begin{tabular}{cccccc}
\textbf{LesionTOADS} & \textbf{S3DL} & \textbf{OASIS} & \textbf{LST} & \textbf{FLEXCONN} & \textbf{Membership} \\  
\includegraphics[width=0.13\textwidth]{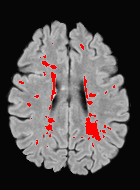} &
\includegraphics[width=0.13\textwidth]{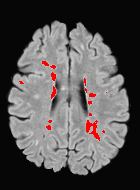} &
\includegraphics[width=0.13\textwidth]{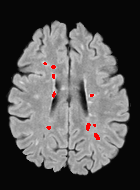} &
\includegraphics[width=0.13\textwidth]{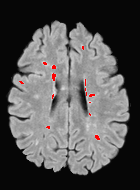} &
\includegraphics[width=0.13\textwidth]{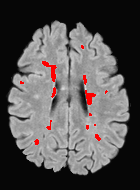} &
\includegraphics[width=0.13\textwidth]{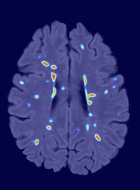} 
\end{tabular}
\tabcolsep 1pt
\begin{tabular}{cccccc}
& \textbf{MPRAGE} & \textbf{FLAIR} & $\mathbf{T_2}$-w & $\mathbf{PD}$-w & \textbf{Manual} \\
{\rotatebox{90}{\hspace{1em}\texttt{Subject \#3}}} &  
\includegraphics[width=0.13\textwidth]{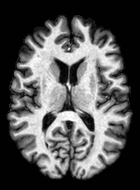} &
\includegraphics[width=0.13\textwidth]{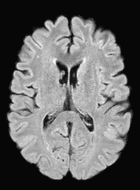} &
\includegraphics[width=0.13\textwidth]{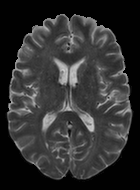} &
\includegraphics[width=0.13\textwidth]{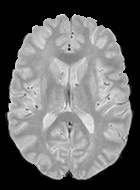} &
\includegraphics[width=0.13\textwidth]{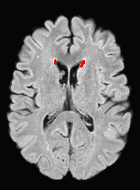} 
\end{tabular}
\tabcolsep 1pt
\begin{tabular}{cccccc}
\textbf{LesionTOADS} & \textbf{S3DL} & \textbf{OASIS} & \textbf{LST} & \textbf{FLEXCONN} & \textbf{Membership} \\  
\includegraphics[width=0.13\textwidth]{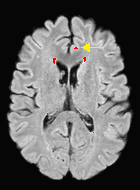} &
\includegraphics[width=0.13\textwidth]{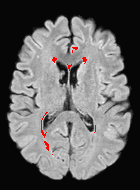} &
\includegraphics[width=0.13\textwidth]{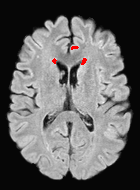} &
\includegraphics[width=0.13\textwidth]{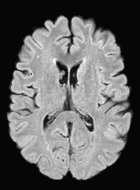} &
\includegraphics[width=0.13\textwidth]{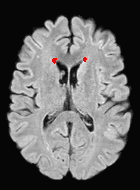} &
\includegraphics[width=0.13\textwidth]{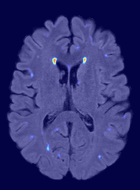} 
\end{tabular}
\end{center}
\vspace{-1em}
\caption{
$T_1$-w, $T_2$-w, $PD$-w, and FLAIR images of three subjects with high, 
medium, and low lesion load from \texttt{MS-100} dataset, along with segmentations from $5$ 
competing methods and one lesion membership from the proposed CNN based FLEXCONN.
}
\label{fig:ms100-example}
\end{figure*}

\begin{figure*}[!tbh]
\begin{center}
\tabcolsep 1pt
\begin{tabular}{cccc}
\textbf{MPRAGE} & \textbf{FLAIR} & \textbf{Rater 1} & \textbf{Rater 2} \\
\includegraphics[width=0.2\textwidth]{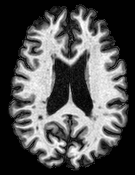} &
\includegraphics[width=0.2\textwidth]{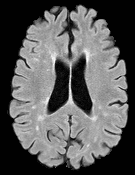} &
\includegraphics[width=0.2\textwidth]{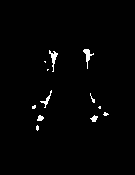} &
\includegraphics[width=0.2\textwidth]{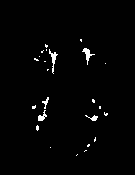} 
\end{tabular}
\begin{tabular}{ccc}
\textbf{Membership\#1} & \textbf{Membership\#2} & \textbf{FLEXCONN}  \\
\includegraphics[width=0.2\textwidth]{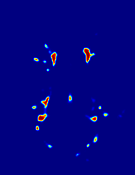} &
\includegraphics[width=0.2\textwidth]{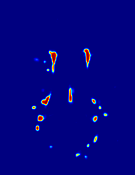} &
\includegraphics[width=0.2\textwidth]{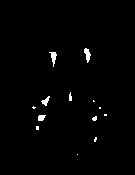} 
\end{tabular}
\end{center}
\caption{A typical segmentation example for one subject from the \texttt{ISBI-61} dataset. 
Membership\#1 and \#2 refer to the lesion memberships obtained using training data from rater 1 and 
2, respectively.
}
\label{fig:isbi-example}
\end{figure*}

\begin{table*}[tbh]
Mean values of comparison metrics are shown for various competing methods on \texttt{ISBI-61} 
dataset. Bold indicates highest or lowest value. See Sec.~\ref{sec:metrics} 
for the definition of the metrics. The score was computed as a weighted average of the other 
metrics.
\begin{center}
\tabcolsep 3pt
\begin{tabular}{cccccc}
\toprule[2pt]
& Dice & LFPR & PPV & VD & Score\\
\cmidrule[1pt](lr){1-6}
Birenbaum \textit{et. al.}\cite{birenbaum2016} & 0.6271 & 0.4976 & 0.7890 & 0.3523 &  90.07\\
Jain      \textit{et. al.}\cite{jain2015}      & 0.5243 & 0.4005 & 0.6947 & 0.3886 &  88.74\\
Tomas-Fernandez \textit{et. al.}\cite{warfield2015}  & 0.4317 & 0.4116 & 0.6974 & 0.5110 &  87.07\\
Gghafoorian \textit{et. al.}\cite{ghafoorian2017a}& 0.5009 & 0.5766 & 0.5942 & 0.5708 &  86.92\\              
Sudre \textit{et. al.}\cite{sudre2015}     & 0.5226 & 0.6776 & 0.6690 & 0.3887 & 86.44\\
Maier \textit{et. al.}\cite{maier2015}     & 0.6050 & 0.2658 & 0.7746 & 0.3654 &  90.28\\
Deshpande \textit{et. al.}\cite{deshpande2015} & 0.5920 & 0.2806 & 0.7622 &  \textbf{0.3214} & 89.81\\
Valverde \textit{et. al.}\cite{llado2017}     & \textbf{0.6305} & 0.1529 & 0.7867 & 0.3385 &  \textbf{91.33} \\
FLEXCONN              & 0.5243 & \textbf{0.1103} & \textbf{0.8660} & 0.5207 & 90.48\\
\bottomrule[1pt]
\end{tabular}
\end{center}
\label{tab:isbi61-volstats}
\end{table*}

Since lesion volume is an important outcome measure for evaluating disease progression, we compared 
automated lesion volume vs the manual lesion volume in Fig.~\ref{fig:ms100-scatterplot}. Solid lines
represent a robust linear fit of the points, and the black dotted line represents unit slope. It is 
observed that LesionTOADS (blue) overestimates lesions when lesion load is small, and LST (magenta)
underestimates the lesion when the lesion load is high. S3DL, OASIS, and FLEXCONN show less bias 
with respect to lesion load, while FLEXCONN has the slope closest to unity ($0.94$). The 
slopes and intercepts with manual lesion volumes are also shown in Table~\ref{tab:ms100-stats}.
Table~\ref{tab:ms100-volstats} shows median values of various comparison metrics for the $5$
competing methods. FLEXCONN produces significantly better Dice, LFPR, and VD ($p<0.01$) 
among the four methods. LST produces the highest PPV.

\subsection{\texttt{ISBI-61} Dataset}
\label{sec:isbi61}
Although \texttt{ISBI-61} includes longitudinal images, we performed the segmentation in a 
cross-sectional manner. The segmentations were generated in a similar fashion as the \texttt{MS-100}
dataset (Sec.~\ref{sec:ms100}) by averaging two memberships obtained using two sets of training.
A typical segmentation example is shown in Fig.~\ref{fig:isbi-example}, where the subject has high 
lesion load ($40$cc).  

Table~\ref{tab:isbi61-volstats} shows a comparison with some of the methods that participated in 
the \texttt{ISBI 2015} challenge. The proposed method achieves the lowest LFPR ($0.1102$) 
and the highest PPV ($0.866$) compared to others, while the highest Dice was produced by another
recent CNN based method \cite{llado2017}. The lowest VD was achieved by a dictionary based 
method \cite{deshpande2015}. To rank a method, a score was computed using a weighted average of 
various metrics including Dice, LFPR, PPV, and VD, as detailed in \cite{carass2017}. For the two 
raters, the inter-rater score was $0.67$, which is scaled to $90$. Therefore, a score of $90$ or 
more indicates segmentation accuracy similar to the consistency between two human raters. FLEXCONN 
achieved a score of $90.48$, while the other CNN based methods, \cite{llado2017} and
\cite{birenbaum2016}, achieved scores of $91.33$ and $90.07$, respectively, indicating their
performance to be comparable to human raters. Most of the top scoring methods in the challenge were 
based on CNN.

\section{Discussion}
We have proposed a simple end-to-end fully convolutional neural network based method to segment 
MS lesions from multi-contrast MR images. Our network is does not have any fully connected layers 
and after training, takes only a couple of seconds to segment lesions on a new image. Although we 
validated using only $T_1$-w and FLAIR contrasts, other contrasts can easily be included in the 
framework. We have shown that using large, two-dimensional, patches provide significantly better 
segmentations than smaller patches. Comparisons with four other publicly available lesion 
segmentation methods, two supervised and two unsupervised, showed superior performance over $100$ 
images.

During training, there were several parameters that were empirically determined. First, for a
$(2w+1)^2$ filter, we used $w^2$ zero padding at each convolution so as to have a uniform input and
output patch size to all filters. Without padding, the output size after every filter bank decreases
and care should be taken to keep the input and output patches properly aligned. With padding, we can
add or remove filter banks without worrying about alignment. Another important parameter is the
batch size. With too small a batch size, the gradient computation becomes noisy and the stochastic
gradient descent optimization may not lead to a local minima. With too large a batch size, the
optimization may lead to a sharp local minima, making the model not generalizable to new data
\cite{keskar2016}.  Therefore, an appropriate batch size should be chosen based on the data. During
training, we empirically chose $128$ for training and $64$ as a test batch size.

With the removal of fully connected layers, the proposed fully convolutional network can generate 
the membership of a 2D slice without the need for dividing images into sub-regions 
\cite{kamnitsas2017}. With large 
enough patches, the contextual information of a lesion voxel can be obtained from within the 
patch. This is representative of a human observer looking at a large neighborhood while considering 
a voxel to be lesion of not. Note that although the training is performed with patches, the 
prediction step does not need the patch information because the trained convolutions are applied to 
a whole 2D slice. As a consequence, the memberships are inherently smooth, and the problem of
possible discontinuities between sub-regions does not arise.

MS lesion segmentation is associated with high inter-rater variability in manual delineations, as
seen on both the \texttt{MICCAI 2008} and \texttt{ISBI 2015} challenges. For
example, in the \texttt{MICCAI 2008} lesion segmentation challenge, the average Dice overlap between
two raters was $0.25$, and in the \texttt{ISBI 2015} challenge, the inter-rater Dice overlap was
$0.63$ \cite{carass2017}. Therefore it is expected that the average Dice coefficients of the
proposed segmentations are as low as $0.5$ and sometimes are even lower. However, Dice coefficients
can be artificially low when the actual lesion volume is small, therefore having fewer false
positives can be more desirable than having a high Dice. Our proposed model had the lowest false
positive rate compared to all other methods on both test datasets while maintaining good
sensitivity.

In our experiments, we used large 2D patches similar to \cite{ghafoorian2017b}, in comparison to 
isotropic 3D patches as used before, e.g., $11^3$ in \cite{llado2017}, $23^3$ in 
\cite{wachinger2017}, and $17^3$ in \cite{kamnitsas2017}. The rationale behind using large 
anisotropic patches is twofold. First, experiments with full 3D isotropic $9^3$ or $11^3$ patches 
showed little or no improvement in Dice and led to increased false positives, with memberships 
similar to the one with $13\times 13$ patches, as shown in
Fig.~\ref{fig:diff_patch_size}. Larger isotropic patches, e.g. $19^3$ or $25^3$, showed inferior
segmentation, and in some cases, optimization did not converge. The reason is that the FLAIR
images in the test datasets had inherently low resolution in the inferior-superior direction,
$2.2$ mm and $4.4$ mm compared to in-plane resolution of $0.82\times 0.82$ mm. Therefore 2D axial
patches capture the high resolution in-plane information that represents the original thick axial
slices. Second, the lesions are usually focal and small in size, unlike other brain
structures. Therefore a very large isotropic patch around a small lesion can include superfluous
information about the lesion, which can increase the amount of false positives. 
Note that with in more recent studies employing high resolution 3D FLAIR sequences, it is 
trivial to extend the algorithm to accommodate for 3D patches.

One drawback of the proposed method is that it requires a large number of training patches. With the
\texttt{ISBI-21} as training data, there are approximately only $270,000$ training patches. 
Patch
rotation \cite{guizard2015} is a standard data augmentation technique where training patches are
rotated by $45$\degree, $90$\degree, $135$\degree, and $180\degree$ in the axial plane, and added to
the training patch set in addition to the non-rotated patches. Our initial experiments with rotated 
patches on the
\texttt{VAL-28} dataset showed only $1-2$\% increase in average Dice coefficients with rotated
patches at the cost of significantly more memory and training time, indicating that the network is
already sufficiently generalizable with the original training data. Therefore we did not use rotated
patches in the final segmentation. However, further experiments are needed to understand the full
scope of performance improvement with respect to the available training data and other augmentation
techniques, such as patch cropping or adding visually imperceptible jitters to images.

Table~\ref{tab:isbi61-volstats} shows that there is no single method that has the highest 
metrics among the six. This is consistent with previously reported results \cite{carass2017} on
the same \texttt{ISBI-61} data. There are several methods with score more than 
$90$, such as \cite{birenbaum2016,maier2015,llado2017}. \cite{llado2017} produced the highest Dice,
while FLEXCONN produced the highest LFPR and PPV. Both these methods are based on CNN, outperforming 
other traditional machine learning based algorithms. Note that FLEXCONN has a very simple network 
architecture and does not have a longitudinal component like \cite{birenbaum2016} or two-pass 
correction mechanism like \cite{llado2017}. Still it was able to achieve similar overall 
performance. Future work will include further comparison with other CNN based methods such 
as \cite{ghafoorian2017a,llado2017,birenbaum2016}. We will also explore more recent and 
state-of-the-art networks such as \cite{szegedy2015,he2016,he2016b} to achieve better accuracy and 
temporal consistency in segmentations.

\section{Acknowledgements}
Support for this work included funding from the Department of Defense in the Center for Neuroscience
and Regenerative Medicine and intramural research program at NIH and NINDS.
This work was also partially supported by grant from National MS Society RG-1507-05243 and
NIH R01NS082347.

\bibliographystyle{elsarticle-harv}
\bibliography{refs}

\begin{thebibliography}{67}
\expandafter\ifx\csname natexlab\endcsname\relax\def\natexlab#1{#1}\fi
\expandafter\ifx\csname url\endcsname\relax
  \def\url#1{\texttt{#1}}\fi
\expandafter\ifx\csname urlprefix\endcsname\relax\def\urlprefix{URL }\fi

\bibitem[{Avants et~al.(2011)Avants, Tustison, Song, Cook, Klein, and
  Gee}]{avants2011}
Avants, B.~B., Tustison, N.~J., Song, G., Cook, P.~A., Klein, A., Gee, J.~C.,
  2011. A reproducible evaluation of {ANTs} similarity metric performance in
  brain image registration. NeuroImage 54~(3), 2033--–2044.

\bibitem[{Birenbaum and Greenspan(2016)}]{birenbaum2016}
Birenbaum, A., Greenspan, H., 2016. Longitudinal multiple sclerosis lesion
  segmentation using multi-view convolutional neural networks. In: Intl.
  Workshop on Deep Learning in Medical Image Analysis. pp. 58--67.

\bibitem[{Brosch et~al.(2016)Brosch, Tang, Yoo, Li, Traboulsee, and
  Tam}]{brosch2016}
Brosch, T., Tang, L. Y.~W., Yoo, Y., Li, D. K.~B., Traboulsee, A., Tam, R.,
  2016. Deep 3d convolutional encoder networks with shortcuts for multiscale
  feature integration applied to multiple sclerosis lesion segmentation. IEEE
  Trans. Med. Imag. 35~(5), 1229--1239.

\bibitem[{Carass et~al.(2011)Carass, Cuzzocreo, Wheeler, Bazin, Resnick, and
  Prince}]{carass2011}
Carass, A., Cuzzocreo, J., Wheeler, M.~B., Bazin, P.~L., Resnick, S.~M.,
  Prince, J.~L., 2011. Simple paradigm for extra-cerebral tissue removal:
  Algorithm and analysis. NeuroImage 56~(4), 1982--–1992.

\bibitem[{Carass and \textit{et. al.}(2017)}]{carass2017}
Carass, A., \textit{et. al.}, 2017. Longitudinal multiple sclerosis lesion
  segmentation: resource \& challenge. NeuroImage 148, 77–--102.

\bibitem[{Casamitjana et~al.(2016)Casamitjana, Puch, Aduriz, and
  Vilaplana}]{veronica2016}
Casamitjana, A., Puch, S., Aduriz, A., Vilaplana, V., 2016. 3d convolutional
  neural networks for brain tumor segmentation: A comparison of
  multi-resolution architectures. In: Intl. Workshop on Brainlesion: Glioma,
  Multiple Sclerosis, Stroke and Traumatic Brain Injuries. pp. 150--161.

\bibitem[{Chen et~al.(2017)Chen, Dou, Yu, Qin, and Heng}]{chen2017}
Chen, H., Dou, Q., Yu, L., Qin, J., Heng, P.-A., 2017. Voxresnet: Deep
  voxelwise residual networks for brain segmentation from {3D MR} images.
  NeuroImage 00, 00.

\bibitem[{Deshpande et~al.(2015)Deshpande, Maurel, and
  Barillot}]{deshpande2015}
Deshpande, H., Maurel, P., Barillot, C., 2015. Adaptive dictionary learning for
  competitive classification of multiple sclerosis lesions. In: Intl. Symp. on
  Biomed. Imag. (ISBI). pp. 136--139.

\bibitem[{Dworkin et~al.(2016)Dworkin, Sweeney, Schindler, Chahin, Reich, and
  Shinohara}]{sweeney2016}
Dworkin, J.~D., Sweeney, E.~M., Schindler, M.~K., Chahin, S., Reich, D.~S.,
  Shinohara, R.~T., 2016. {PREVAIL}: Predicting recovery through estimation and
  visualization of active and incident lesions. NeuroImage: Clinical 12,
  293--299.

\bibitem[{Egger et~al.(2017)Egger, Opfer, Wang, Kepp, Sormani, Spies, Barnett,
  and Schippling}]{egger2017}
Egger, C., Opfer, R., Wang, C., Kepp, T., Sormani, M.~P., Spies, L., Barnett,
  M., Schippling, S., 2017. {MRI FLAIR} lesion segmentation in multiple
  sclerosis: Does automated segmentation hold up with manual annotation?
  NeuroImage 13, 264--270.

\bibitem[{Freire and Ferrari(2016)}]{ferrari2016}
Freire, P.~G., Ferrari, R.~J., 2016. Automatic iterative segmentation of
  multiple sclerosis lesions using {Student's} t mixture models and
  probabilistic anatomical atlases in {FLAIR} images. Computers in Biology and
  Medicine 73, 10--23.

\bibitem[{Garcia-Lorenzo et~al.(2013)Garcia-Lorenzo, Francis, Narayanan,
  Arnold, and Collins}]{lorenzo2013}
Garcia-Lorenzo, D., Francis, S., Narayanan, S., Arnold, D.~L., Collins, D.~L.,
  2013. Review of automatic segmentation methods of multiple sclerosis white
  matter lesions on conventional magnetic resonance imaging. Med. Image Anal.
  17~(1), 1–--18.

\bibitem[{Garcia-Lorenzo et~al.(2011)Garcia-Lorenzo, Prima, Arnold, Collins,
  and Barillot}]{lorenzo2011}
Garcia-Lorenzo, D., Prima, S., Arnold, D.~L., Collins, L.~D., Barillot, C.,
  2011. Trimmed-likelihood estimation for focal lesions and tissue segmentation
  in multisequence {MRI} for multiple sclerosis. IEEE Trans. Med. Imag. 30~(8),
  1455–--1467.

\bibitem[{Geremia et~al.(2011)Geremia, Clatz, Menze, Konukoglu, Criminisi, and
  Ayache}]{geremia2011}
Geremia, E., Clatz, O., Menze, B.~H., Konukoglu, E., Criminisi, A., Ayache, N.,
  2011. Spatial decision forests for {MS} lesion segmentation in multi-channel
  magnetic resonance images. NeuroImage 57~(2), 378--390.

\bibitem[{Ghafoorian et~al.(2017{\natexlab{a}})Ghafoorian, Karssemeijer,
  Heskes, Bergkamp, Wissink, Obels, Keizer, {de Leeuw}, {van Ginneken},
  Marchiori, and Platel}]{ghafoorian2017a}
Ghafoorian, M., Karssemeijer, N., Heskes, T., Bergkamp, M., Wissink, J., Obels,
  J., Keizer, K., {de Leeuw}, F.-E., {van Ginneken}, B., Marchiori, E., Platel,
  B., 2017{\natexlab{a}}. Deep multi-scale location-aware {3D} convolutional
  neural networks for automated detection of lacunes of presumed vascular
  origin. NeuroImage: Clinical 14, 391--399.

\bibitem[{Ghafoorian et~al.(2017{\natexlab{b}})Ghafoorian, Karssemeijer,
  Heskes, {van Uden}, Sanchez, Litjens, {de Leeuw}, {van Ginneken}, Marchiori,
  and Platel}]{ghafoorian2017b}
Ghafoorian, M., Karssemeijer, N., Heskes, T., {van Uden}, I. W.~M., Sanchez,
  C.~I., Litjens, G., {de Leeuw}, F.~E., {van Ginneken}, B., Marchiori, E.,
  Platel, B., 2017{\natexlab{b}}. Location sensitive deep convolutional neural
  networks for segmentation of white matter hyperintensities. Scientific
  Reports 7, 5110.

\bibitem[{Greenspan et~al.(2016)Greenspan, van Ginneken, and
  Summers}]{summers2016}
Greenspan, H., van Ginneken, B., Summers, R.~M., 2016. Guest editorial deep
  learning in medical imaging: Overview and future promise of an exciting new
  technique. IEEE Trans. Med. Imag. 35~(5), 1153--1159.

\bibitem[{Griffanti et~al.(2016)Griffanti, Zamboni, Khan, Li, Bonifacio,
  Sundaresan, Schulz, Kuker, Battaglini, and Rothwell}]{griffanti2016}
Griffanti, L., Zamboni, G., Khan, A., Li, L., Bonifacio, G., Sundaresan, V.,
  Schulz, U.~G., Kuker, W., Battaglini, M., Rothwell, P.~M., 2016. {BIANCA}
  ({Brain Intensity AbNormality Classification Algorithm}): A new tool for
  automated segmentation of white matter hyperintensities. NeuroImage 141,
  191–205.

\bibitem[{Guizard et~al.(2015)Guizard, Coupe, Fonov, Manjon, Arnold, and
  Collins}]{guizard2015}
Guizard, N., Coupe, P., Fonov, V.~S., Manjon, J.~V., Arnold, D.~L., Collins,
  D.~L., 2015. Rotation-invariant multi-contrast non-local means for {MS}
  lesion segmentation. NeuroImage: Clinical 8, 376–--389.

\bibitem[{Harmouche et~al.(2006)Harmouche, Collins, Arnold, Francis, and
  Arbel}]{harmouche2006}
Harmouche, R., Collins, L., Arnold, D., Francis, S., Arbel, T., 2006. {Bayesian
  MS Lesion Classification Modeling Regional and Local Spatial Information}.
  IEEE Intl. Conf. Patt. Recog. 3, 984--987.

\bibitem[{Harmouche et~al.(2015)Harmouche, Subbanna, Collins, Arnold, and
  Arbel}]{harmouche2015}
Harmouche, R., Subbanna, N.~K., Collins, D.~L., Arnold, D.~L., Arbel, T., 2015.
  Probabilistic multiple sclerosis lesion classification based on modeling
  regional intensity variability and local neighborhood information. IEEE
  Trans. Biomed. Engg. 62~(5), 1281--1292.

\bibitem[{Havaei et~al.(2016)Havaei, Guizard, Chapados, and
  Bengio}]{bengio2016}
Havaei, M., Guizard, N., Chapados, N., Bengio, Y., 2016. {HeMIS}: Hetero-modal
  image segmentation. In: Med. Image Comp. and Comp. Asst. Intervention
  (MICCAI). pp. 469--477.

\bibitem[{He and Garcia(2009)}]{he2009}
He, H., Garcia, E.~A., 2009. Learning from imbalanced data. IEEE Trans.
  Knowledge and Data Engineering 21~(9), 1263--1284.

\bibitem[{He et~al.(2016{\natexlab{a}})He, Zhang, Ren, and Sun}]{he2016}
He, K., Zhang, X., Ren, S., Sun, J., 2016{\natexlab{a}}. Deep residual learning
  for image recognition. In: Intl. Conf. on Comp. Vision. and Patt. Recog.
  (CVPR). pp. 770--778.

\bibitem[{He et~al.(2016{\natexlab{b}})He, Zhang, Ren, and Sun}]{he2016b}
He, K., Zhang, X., Ren, S., Sun, J., 2016{\natexlab{b}}. Identity mappings in
  deep residual networks. In: European Conf. on Comp. Vision (ECCV). pp.
  630--645.

\bibitem[{Jain et~al.(2015)Jain, Sima, Ribbens, Cambron, Maertens, Hecke, Mey,
  Barkhof, Steenwijk, Daams, Maes, Huffel, Vrenken, and Smeets}]{jain2015}
Jain, S., Sima, D.~M., Ribbens, A., Cambron, M., Maertens, A., Hecke, W.~V.,
  Mey, J.~D., Barkhof, F., Steenwijk, M.~D., Daams, M., Maes, F., Huffel,
  S.~V., Vrenken, H., Smeets, D., 2015. Automatic segmentation and volumetry of
  multiple sclerosis brain lesions from {MR} images. NeuroImage: Clinical
  8~(5), 1229--1239.

\bibitem[{Jog et~al.(2015)Jog, Carass, Pham, and Prince}]{jog2015}
Jog, A., Carass, A., Pham, D.~L., Prince, J.~L., 2015. Multi-output decision
  trees for lesion segmentation in multiple sclerosis. In: Proceedings of SPIE
  Medical Imaging (SPIE). Vol. 9413. p. 94131C.

\bibitem[{Kalincik et~al.(2012)Kalincik, Vaneckova, Tyblova, Krasensky, Seidl,
  Havrdova, and Horakova}]{kalincik2012}
Kalincik, T., Vaneckova, M., Tyblova, M., Krasensky, J., Seidl, Z., Havrdova,
  E., Horakova, D., 2012. Volumetric {MRI} markers and predictors of disease
  activity in early multiple sclerosis: A longitudinal cohort study. PLoS One
  7~(11), e50101.

\bibitem[{Kamnitsas et~al.(2017)Kamnitsas, Ledig, Newcombe, Simpson, Kane,
  Menon, Rueckert, and Glocker}]{kamnitsas2017}
Kamnitsas, K., Ledig, C., Newcombe, V.~F., Simpson, J.~P., Kane, A.~D., Menon,
  D.~K., Rueckert, D., Glocker, B., 2017. Efficient multi-scale {3D CNN} with
  fully connected {CRF} for accurate brain lesion segmentation. Med. Image
  Anal. 36, 61--–78.

\bibitem[{Keskar et~al.(2016)Keskar, Mudigere, Nocedal, Smelyanskiy, and
  Tang}]{keskar2016}
Keskar, N.~S., Mudigere, D., Nocedal, J., Smelyanskiy, M., Tang, P. T.~P.,
  2016. On large-batch training for deep learning: Generalization gap and sharp
  minima. arXiv preprint arXiv:1609.04836.

\bibitem[{Kingma and Ba(2015)}]{kingma2015}
Kingma, D.~P., Ba, J., 2015. Adam: A method for stochastic optimization. In:
  Intl. Conf. on Learning Representations (ICLR).

\bibitem[{Kleesiek et~al.(2016)Kleesiek, Urban, Hubert, Schwarz, Maier-Hein,
  Bendszus, and Biller}]{kleesiek2016}
Kleesiek, J., Urban, G., Hubert, A., Schwarz, D., Maier-Hein, K., Bendszus, M.,
  Biller, A., 2016. Deep {MRI} brain extraction: A {3D} convolutional neural
  network for skull stripping. NeuroImage 129, 460--469.

\bibitem[{Lao et~al.(2008)Lao, Shen, Liu, Jawad, Melhem, Launer, Bryan, and
  Davatzikos}]{christos2008}
Lao, Z., Shen, D., Liu, D., Jawad, A.~F., Melhem, E.~R., Launer, L.~J., Bryan,
  R.~N., Davatzikos, C., 2008. Computer-assisted segmentation of white matter
  lesions in {3D MR} images, using support vector machine. Academic Radiology
  15~(3), 300--313.

\bibitem[{LeCun et~al.(2015)LeCun, Bengio, and Hinton}]{hinton2015}
LeCun, Y., Bengio, Y., Hinton, G., 2015. Deep learning. Nature 521~(7553),
  436–444.

\bibitem[{Leemput et~al.(2001)Leemput, Maes, Vandermeulen, Colchester, and
  Suetens}]{leemput2001}
Leemput, K.~V., Maes, F., Vandermeulen, D., Colchester, A., Suetens, P., 2001.
  Automated segmentation of multiple sclerosis lesions by model outlier
  detection. IEEE Trans. Med. Imag. 20~(8), 677--688.

\bibitem[{Litjens et~al.(2017)Litjens, Kooi, Bejnordi, Setio, Ciompi,
  Ghafoorian, {van der Laak}, {van Ginneken}, and Sanchez}]{litjens2017}
Litjens, G., Kooi, T., Bejnordi, B.~E., Setio, A. A.~A., Ciompi, F.,
  Ghafoorian, M., {van der Laak}, J.~A., {van Ginneken}, B., Sanchez, C.~I.,
  2017. A survey on deep learning in medical image analysis. Med. Image Anal.
  42, 60--88.

\bibitem[{Maier et~al.(2015)Maier, Wilms, {von der Gablentz}, Kramer, Munte,
  and Handels}]{maier2015}
Maier, O., Wilms, M., {von der Gablentz}, J., Kramer, U.~M., Munte, T.~F.,
  Handels, H., 2015. Extra tree forests for sub-acute ischemic stroke lesion
  segmentation in {MR} sequences. Journal of Neuroscience Methods 89, 89--100.

\bibitem[{Menze and \textit{et. al.}(2015)}]{menze2015}
Menze, B.~H., \textit{et. al.}, 2015. The multimodal brain tumor image
  segmentation benchmark {(BRATS)}. IEEE Trans. Med. Imag. 34~(10),
  1993--–2024.

\bibitem[{Moeskops et~al.(2017)Moeskops, {de Bresser}, Kuijf, Mendrik,
  Biessels, Pluim, and Isgum}]{moeskops2017}
Moeskops, P., {de Bresser}, J., Kuijf, H., Mendrik, A., Biessels, G., Pluim, J.
  P.~W., Isgum, I., 2017. Evaluation of a deep learning approach for the
  segmentation of brain tissues and white matter hyperintensities of presumed
  vascular origin in {MRI}. NeuroImage: Clinical 00, 00--00.

\bibitem[{Moeskops et~al.(2016{\natexlab{a}})Moeskops, Viergever, Mendrik, {de
  Vries}, Benders, and Isgum}]{isgum2016b}
Moeskops, P., Viergever, M.~A., Mendrik, A.~M., {de Vries}, L.~S., Benders, M.
  J. N.~L., Isgum, I., 2016{\natexlab{a}}. Automatic segmentation of {MR} brain
  images with a convolutional neural network. IEEE Trans. Med. Imag. 35~(5),
  1252--1261.

\bibitem[{Moeskops et~al.(2016{\natexlab{b}})Moeskops, Wolterink, Velden,
  Gilhuijs, Leiner, Viergever, and Isgum}]{isgum2016a}
Moeskops, P., Wolterink, J.~M., Velden, B. H. M.~v., Gilhuijs, K. G.~A.,
  Leiner, T., Viergever, M.~A., Isgum, I., 2016{\natexlab{b}}. Deep learning
  for multi-task medical image segmentation in multiple modalities. In: Med.
  Image Comp. and Comp. Asst. Intervention (MICCAI). pp. 478--486.

\bibitem[{Nair and Hinton(2010)}]{hinton2010}
Nair, V., Hinton, G.~E., 2010. Rectified linear units improve restricted
  boltzmann machines. In: Intl. Conf. on Machine Learning (ICML). pp. 807--814.

\bibitem[{Oishi et~al.(2008)Oishi, Zilles, Amunts, Faria, Jiang, Li, Akhter,
  Hua, Woods, Toga, Pike, Rosa-Neto, Evans, Zhang, Huang, Miller, {van Zijl},
  Mazziotta, and Mori}]{mori2009}
Oishi, K., Zilles, K., Amunts, K., Faria, A., Jiang, H., Li, X., Akhter, K.,
  Hua, K., Woods, R., Toga, A.~W., Pike, G.~B., Rosa-Neto, P., Evans, A.,
  Zhang, J., Huang, H., Miller, M.~I., {van Zijl}, P.~C., Mazziotta, J., Mori,
  S., 2008. Human brain white matter atlas: identification and assignment of
  common anatomical structures in superficial white matter. NeuroImage 43~(3),
  447--457.

\bibitem[{Pereira et~al.(2016)Pereira, Pinto, Alves, and Silva}]{pereira2015}
Pereira, S., Pinto, A., Alves, V., Silva, C.~A., 2016. Brain tumor segmentation
  using convolutional neural networks in {MRI} images. IEEE Trans. Med. Imag.
  35~(5), 1240--1251.

\bibitem[{Prieto et~al.(2017)Prieto, Cavallari, Palotai, Pinzon, Egorova,
  Styner, and Guttmann}]{prieto2017}
Prieto, J.~C., Cavallari, M., Palotai, M., Pinzon, A.~M., Egorova, S., Styner,
  M., Guttmann, C. R.~G., 2017. Large deep neural networks for {MS} lesion
  segmentation. In: Proceedings of SPIE Medical Imaging (SPIE). Vol. 10133. p.
  10133F.

\bibitem[{Roth et~al.(2016)Roth, Lu, Liu, Yao, Seff, Cherry, Kim, and
  Summers}]{roth2016}
Roth, H.~R., Lu, L., Liu, J., Yao, J., Seff, A., Cherry, K., Kim, L., Summers,
  R.~M., 2016. Improving computer-aided detection using convolutional neural
  networks and random view aggregation. IEEE Trans. Med. Imag. 35~(5),
  1170--1181.

\bibitem[{Roura et~al.(2015)Roura, Oliver, Cabezas, Valverde, Pareto, Vilanova,
  Ramio-Torrenta, Rovira, and Llado}]{llado2015}
Roura, E., Oliver, A., Cabezas, M., Valverde, S., Pareto, D., Vilanova, J.~C.,
  Ramio-Torrenta, L., Rovira, A., Llado, X., 2015. A toolbox for multiple
  sclerosis lesion segmentation. Neuroradiology 57, 1031--–1043.

\bibitem[{Roy et~al.(2017)Roy, Butman, Pham, and {Alzheimers Disease
  Neuroimaging Initiative}}]{roy2017}
Roy, S., Butman, J.~A., Pham, D.~L., {Alzheimers Disease Neuroimaging
  Initiative}, 2017. Robust skull stripping using multiple {MR} image contrasts
  insensitive to pathology. NeuroImage 146, 132–--147.

\bibitem[{Roy et~al.(2015{\natexlab{a}})Roy, Carass, Prince, and
  Pham}]{roy2015mlmi}
Roy, S., Carass, A., Prince, J.~L., Pham, D.~L., 2015{\natexlab{a}}.
  Longitudinal patch-based segmentation of multiple sclerosis white matter
  lesions. In: Machine Learning in Medical Imaging. Vol. 9352. pp. 194--202.

\bibitem[{Roy et~al.(2014)Roy, He, Carass, Jog, Cuzzocreo, Reich, Prince, and
  Pham}]{roy2014spie1}
Roy, S., He, Q., Carass, A., Jog, A., Cuzzocreo, J.~L., Reich, D.~S., Prince,
  J.~L., Pham, D.~L., 2014. Example based lesion segmentation. In: Proceedings
  of SPIE Medical Imaging (SPIE). Vol. 9034. p. 90341Y.

\bibitem[{Roy et~al.(2015{\natexlab{b}})Roy, He, Sweeney, Carass, Reich,
  Prince, and Pham}]{roy2015}
Roy, S., He, Q., Sweeney, E., Carass, A., Reich, D.~S., Prince, J.~L., Pham,
  D.~L., 2015{\natexlab{b}}. Subject specific sparse dictionary learning for
  atlas based brain {MRI} segmentation. IEEE Journal of Biomedical and Health
  Informatics 19~(5), 1598--1609.

\bibitem[{Sati and \textit{et. al.}(2016)}]{sati2016}
Sati, P., \textit{et. al.}, 2016. The central vein sign and its clinical
  evaluation for the diagnosis of multiple sclerosis: a consensus statement
  from the {North American Imaging in Multiple Sclerosis Cooperative}. Nature
  Rev. Neurology 12, 714–--722.

\bibitem[{Schmidt et~al.(2012)Schmidt, Gaser, Arsic, Buck, Forschler, Berthele,
  Hoshi, Ilg, Schmid, Zimmer, Hemmer, and Muhlau}]{schmidt2012}
Schmidt, P., Gaser, C., Arsic, M., Buck, D., Forschler, A., Berthele, A.,
  Hoshi, M., Ilg, R., Schmid, V.~J., Zimmer, C., Hemmer, B., Muhlau, M., 2012.
  An automated tool for detection of {FLAIR}-hyperintense white-matter lesions
  in multiple sclerosis. NeuroImage 59~(4), 3774--3783.

\bibitem[{Shiee et~al.(2009)Shiee, Bazin, Ozturk, Reich, Calabresi, and
  Pham}]{shiee2009}
Shiee, N., Bazin, P.~L., Ozturk, A., Reich, D.~S., Calabresi, P.~A., Pham,
  D.~L., 2009. {A Topology-Preserving Approach to the Segmentation of Brain
  Images with Multiple Sclerosis Lesions}. NeuroImage 49~(2), 1524--–1535.

\bibitem[{Simonyan and Zisserman(2015)}]{karen2015}
Simonyan, K., Zisserman, A., 2015. Very deep convolutional networks for
  large-scale image recognition. In: Intl. Conf. on Learning Representations
  (ICLR).

\bibitem[{Souplet et~al.(2008)Souplet, Lebrun, Ayache, and
  Malandain}]{souplet2008}
Souplet, J., Lebrun, C., Ayache, N., Malandain, G., 2008. An automatic
  segmentation of {T2-FLAIR} multiple sclerosis lesions. In: Multiple Sclerosis
  Lesion Segmentation Challenge Workshop (MICCAI 2008 Workshop).

\bibitem[{Srivastava et~al.(2014)Srivastava, Hinton, Krizhevsky, Sutskever, and
  Salakhutdinov}]{srivastava2014}
Srivastava, N., Hinton, G., Krizhevsky, A., Sutskever, I., Salakhutdinov, R.,
  2014. Dropout: A simple way to prevent neural networks from overfitting. J.
  Machine Learning Research 15~(1), 1929--1958.

\bibitem[{Strumia et~al.(2016)Strumia, Schmidt, Anastasopoulos, Granziera,
  Krueger, and Brox}]{strumia2016}
Strumia, M., Schmidt, F.~R., Anastasopoulos, C., Granziera, C., Krueger, G.,
  Brox, T., 2016. White matter {MS}-lesion segmentation using a geometric brain
  model. IEEE Trans. Med. Imag. 35~(2), 1636--1646.

\bibitem[{Sudre et~al.(2015)Sudre, Cardoso, Bouvy, Biessels, Barnes, and
  Ourselin}]{sudre2015}
Sudre, C.~H., Cardoso, M.~J., Bouvy, W.~H., Biessels, G.~J., Barnes, J.,
  Ourselin, S., 2015. Bayesian model selection for pathological neuroimaging
  data applied to white matter lesion segmentation. IEEE Trans. Med. Imag.
  34~(10), 2079--2102.

\bibitem[{Sweeney et~al.(2013)Sweeney, Shinohara, Shiee, Mateen, Chudgar,
  Cuzzocreo, Calabresi, Pham, and Reich}]{sweeney2013}
Sweeney, E.~M., Shinohara, R.~T., Shiee, N., Mateen, F.~J., Chudgar, A.~A.,
  Cuzzocreo, J.~L., Calabresi, P.~A., Pham, D.~L., Reich, D.~S., 2013. {OASIS}
  is automated statistical inference for segmentation, with applications to
  multiple sclerosis lesion segmentation in {MRI}. NeuroImage: Clinical 2,
  402–413.

\bibitem[{Szegedy et~al.(2015)Szegedy, Liu, Jia, Sermanet, Reed, Anguelov,
  Erhan, Vanhoucke, and Rabinovich}]{szegedy2015}
Szegedy, C., Liu, W., Jia, Y., Sermanet, P., Reed, S., Anguelov, D., Erhan, D.,
  Vanhoucke, V., Rabinovich, A., 2015. Going deeper with convolutions. In:
  Intl. Conf. on Comp. Vision. and Patt. Recog. (CVPR). pp. 1--9.

\bibitem[{Tomas-Fernandez and Warfield(2015)}]{warfield2015}
Tomas-Fernandez, X., Warfield, S.~K., 2015. A model of population and subject
  {(MOPS)} intensities with application to multiple sclerosis lesion
  segmentation. IEEE Trans. Med. Imag. 34~(6), 1349--1361.

\bibitem[{Tustison et~al.(2010)Tustison, Avants, Cook, Zheng, Egan, Yushkevich,
  and Gee}]{tustison2010}
Tustison, N.~J., Avants, B.~B., Cook, P.~A., Zheng, Y., Egan, A., Yushkevich,
  P.~A., Gee, J.~C., 2010. {N4ITK}: improved {N3} bias correction. IEEE Trans.
  Med. Imag. 29~(6), 1310--1320.

\bibitem[{Valverde et~al.(2017)Valverde, Cabezasa, Roura, González-Villaa,
  Pareto, Vilanova, Ramio-Torrenta, Rovira, Oliver, and Llado}]{llado2017}
Valverde, S., Cabezasa, M., Roura, E., González-Villaa, S., Pareto, D.,
  Vilanova, J.~C., Ramio-Torrenta, L., Rovira, A., Oliver, A., Llado, X., 2017.
  Improving automated multiple sclerosis lesion segmentation with a cascaded 3d
  convolutional neural network approach. NeuroImage 155, 159–--168.

\bibitem[{Wachinger et~al.(2017)Wachinger, Reuter, and Klein}]{wachinger2017}
Wachinger, C., Reuter, M., Klein, T., 2017. {DeepNAT}: Deep convolutional
  neural network for segmenting neuroanatomy. NeuroImage 00~(00), 00.

\bibitem[{Yoo et~al.(2014)Yoo, Brosch, Traboulsee, Li, and Tam}]{yoo2014}
Yoo, Y., Brosch, T., Traboulsee, A., Li, D.~K., Tam, R., 2014. Deep learning of
  image features from unlabeled data for multiple sclerosis lesion
  segmentation. In: Machine Learning in Medical Imaging. Vol. 8679. pp.
  117–--124.

\bibitem[{Zhang et~al.(2015)Zhang, Li, Deng, Wang, Lin, Ji, and
  Shen}]{zhang2015}
Zhang, W., Li, R., Deng, H., Wang, L., Lin, W., Ji, S., Shen, D., 2015. Deep
  convolutional neural networks for multi-modality isointense infant brain
  image segmentation. NeuroImage 108, 214–224.

\end{thebibliography}


\end{document}